\documentclass[preprint,11pt,sort&compress]{elsarticle}
\usepackage[margin=2.5cm]{geometry}
\usepackage[english]{babel} 
\usepackage{graphicx}
\usepackage[hidelinks]{hyperref}
\usepackage{url}
\usepackage{amsmath}
\usepackage{float}
\usepackage{amsfonts}
\usepackage{cleveref}
\usepackage{algorithm}
\usepackage{algpseudocode}
\usepackage[modulo]{lineno}
\usepackage{circuitikz}
\usepackage[edges]{forest}
\usepackage{array}
\usetikzlibrary{shapes.gates.logic.US,trees,positioning,arrows,shapes.geometric}
\usetikzlibrary{calc,patterns,trees,decorations.pathmorphing,decorations.markings}
\usepackage{multicol}
\usepackage{rotating}
\usepackage{tikz}
\usepackage{pgfplots}
\usetikzlibrary{bayesnet}
\usetikzlibrary{arrows}
\usepackage{subcaption}

\usepackage{orcidlink}
\usepackage{xparse}
\usepackage{mathtools}
\usepackage{amsthm,amsmath,amssymb}

\renewcommand{\epsilon}{\varepsilon} 
\newcommand{\elmtreal}[1]{\in \mathbb{R}^{#1}}

\renewcommand{\vec}[1]{\boldsymbol{\mathbf{#1}}} 
\newcommand{\mat}[1]{\vec{\MakeUppercase{#1}}} 
\newcommand{\trace}[1]{\mathrm{Tr}\left(#1\right)} 
\newcommand{\ident}{\mathbb{I}} 
\newcommand{\trans}{{\mathsf{T}}} 
\newcommand{\hlf}{\frac{1}{2}} 

\newcommand{\expect}[2][]{\mathbb{E}_{#1}\left[#2\right]}

\newcommand{\covar}[1]{\mat{\Sigma}^{#1}}

\newcommand{\argmax}[1]{\underset{#1}{\text{arg\,max}}}

\newcommand{\mean}[1]{\vec{\mu}^{#1}}

\newcommand{\gaussdistr}[2]{\mathcal{N}\left(#1\,,#2\right)} 
\newcommand{\iwishdistr}[2]{I\mathcal{W}\left(#1\,,#2\right)} 
\newcommand{\wishdistr}[2]{\mathcal{W}\left(#1\,,#2\right)} 

\newcommand{\wght}[1]{\mat{W}^{#1}}

\newcommand{\wghtmle}[1]{\mat{\hat{W}}^{#1}}

\newcommand{\xstar}[1][]{
\ifthenelse{\equal{#1}{vec}}{\mat{X}_\star}{\vec{x}_\star}
}
\newcommand{\fstar}[1][]{
\ifthenelse{\equal{#1}{vec}}{\vec{f}_\star}{f_\star}
}
\newcommand{\ystar}[1][]{
\ifthenelse{\equal{#1}{vec}}{\vec{y}_\star}{y_\star}
}
\newcommand{\ypred}[1][]{
\ifthenelse{\equal{#1}{vec}}{\hat{\vec{y}}}{\hat{y}}
}
\newcommand{\ytrue}[1][]{
\ifthenelse{\equal{#1}{vec}}{\vec{y}}{y}
}

\journal{Mechanical Systems and Signal processing (MSSP)}

\begin{document}

\begin{frontmatter}

	\title{A new perspective on Bayesian Operational Modal Analysis} 

	\author[1]{Brandon J. O'Connell \orcidlink{0000-0001-6042-927X} \corref{cor1}}
	\ead{b.j.oconnell@sheffield.ac.uk}
	\address[1]{Dynamics Research Group, Department of Mechanical Engineering, The University of Sheffield, Mappin Street, Sheffield, S1 3JD, UK}
	\cortext[cor1]{Corresponding Author}
	\author[1]{Max D. Champneys \orcidlink{0000-0002-3037-7584}}
	\author[1]{Timothy J. Rogers \orcidlink{0000-0002-3433-3247}}

\begin{abstract}
	The quantification of uncertainty is of particular interest to the dynamics community, which increasingly desires a measure of uncertainty for greater insight, allowing for more informed and confident decision-making. In the field of operational modal analysis (OMA), obtained modal information is frequently used to assess the current state of aerospace, mechanical, offshore and civil structures. However, the stochasticity of operational systems and the lack of forcing information can lead to inconsistent results. Quantifying the uncertainty of the recovered modal parameters through OMA is therefore of significant value. In this article, a new perspective on Bayesian OMA is proposed --- a Bayesian stochastic subspace identification (SSI) algorithm. Distinct from existing approaches to Bayesian OMA, a hierarchical probabilistic model is embedded at the core of canonical variate-weighted, covariance-driven SSI. Through substitution of canonical correlation analysis with a Bayesian equivalent, posterior distributions over the modal characteristics are obtained. Two inference schemes are presented for the proposed Bayesian formulation: Markov Chain Monte Carlo and variational Bayes. Two case studies are then explored. The first is benchmark study using data from a simulated, multi degree-of-freedom, linear system. Following application of Bayesian SSI using both forms of inference, it is shown that the same posterior is targeted and recovered by both schemes, with good agreement between the mean of the posterior and the conventional SSI result. The second study applies the variational form of Bayesian SSI to data obtained from an in-service structure --- the Z24 bridge. The Z24 is chosen given its familiarity in the fields of OMA and structural health monitoring. The results of this study are first presented at a single model order, and then at multiple model orders using a stabilisation diagram. In both cases, the recovered posterior uncertainty is included and compared to the conventional SSI result. It is observed that the posterior distributions with mean values coinciding with the natural frequencies exhibit much lower variance than posteriors situated away from the natural frequencies. 
	
\end{abstract}
	

\end{frontmatter}

	
\section{Introduction}\label{sec:Introduction}

The characterisation of dynamical systems in the absence of measured input information continues to be of significant importance in modern engineering practice. This problem is of particular interest to the structural dynamics community, who routinely encounter such scenarios when conducting modal analysis. Operational modal analysis (OMA) is the subset of modal analysis methods concerned with the recovery of modal characteristics in the absence of measured input information; often the case when testing a structure in-situ (operationally) \cite{brincker2015}. By the very nature of OMA, response data is typically obtained at very low amplitudes lending itself to low signal-to-noise ratios. This tends to make the task of system identification more challenging. Nevertheless, various approaches to OMA exist in the literature and work well in practice \cite{brincker2015}. These include methodologies such as frequency-domain decomposition (FDD) \cite{brincker2001} and stochastic subspace identification (SSI) \cite{vanoverschee1996,katayama2005}, now industry standards for performing frequency-domain and time-domain OMA. OMA as a methodology has gained increased popularity in recent years. This is predominantly due to its high economic value and convenience in many engineering applications. This is especially the case for high-value large-scale assets \cite{brincker2015}.

With increasing amounts of data being measured and growing demand for improved data-based models, there is now a desire in the engineering community to obtain some measure of the uncertainty. Access to uncertainty can allow the practitioner to better assess the precision of a chosen methodology and evaluate the risk of different outcomes. The inclusion and application of uncertainty through probabilistic machine learning and Bayesian methodologies has already been observed in the closely related fields of structural health monitoring (SHM) \cite{farrar2013, mclean2023, li2016, bull2021,miah2022}, digital-twins \cite{wagg2020a,thelen2023,rios2020,kochunas2021} and risk-based decision-making \cite{hughes2022}. From the perspective of OMA, understanding and assessing the uncertainty of the recovered modal characteristics, i.e. the natural frequencies, damping ratios and mode shapes, is of particular value.  Multiple sources of uncertainty can result in variability in the recovered modal parameters. Understanding this variability is important in SHM for example, where modal information is often used to assess the health state of a structure. The impacts of aleatory and epistemic uncertainties on operationally obtained modal identification has been studied in \cite{ciloglu2012}, whilst a comprehensive list of the associated uncertainties for SSI are described in \cite{reynders2008}. Fundamentally, being able to quantify the uncertainty over the modal properties can provide the practitioner with greater insight and additional information, allowing for more-informed and confident decision-making. Furthermore, the authors believe probabilistic system identification tools and the inclusion of uncertainty may provide a framework for addressing many of the current research challenges in OMA.

Across different methodologies, the definition of `uncertainty' can vary, taking many forms: bounds, confidence intervals, fiducial intervals, variance estimates, and distributions. There is no explicit definition for the correct type of uncertainty in any case, however the overall objective of uncertainty quantification (UQ) is the same: To obtain a measure of the uncertainty to better understand the variability of a given result. The authors' preference is to operate within a Bayesian framework where the recovery of \emph{posterior distributions} is desired, hence the focus of this paper.

\subsection{Uncertainty quantification for OMA}

In the literature, UQ for OMA is approached and handled in a variety of ways. Several methodologies for quantifying uncertainty in output-only testing have been presented over the last 30 years, taking a variety of approaches including autoregressive techniques \cite{yang2019,kang2023}, Bayesian methods \cite{rogers2020a,au2012,au2012a,beck2010a,huang2017}, sensitivity analysis and perturbation theory \cite{reynders2016,reynders2008,reynders2021,dohler2013a,dohler2013,gres2022,gres2022a}.\\

One of the earliest works on this topic, by Pintelon et.al. \cite{pintelon2007}, saw the derivation of expressions for the uncertainty bounds on the estimated modal parameters using a combination of first-order sensitivity and perturbation techniques. This method was shown specifically for reference-based SSI by Reynders et.al in \cite{reynders2008} where covariances on the identified system matrices were recovered. Later this method was efficiently optimised through rederivation by D\"{o}hler \cite{dohler2013a} and also extended for multiple-setup measurements in \cite{dohler2013}. Other notable works include that of El-Kafafy et.al. \cite{el-kafafy2013a}, who presented a fast maximum-likelihood identification method for obtaining modal parameters with uncertainty intervals, and that of Mellinger et.al. \cite{mellinger2016} who employ sensitivity analysis of the auto- and cross-covariance matrices to obtain variances over the modal parameters. 

Many recent developments in UQ for OMA have centred around subspace methods, like SSI. This is predominantly because of their praised performance in OMA scenarios; demonstrating high estimation accuracy coupled with high computational robustness and overall efficiency. Several notable advancements in SSI-based UQ methods have been presented by a collection of authors including Reynders \cite{reynders2008,reynders2012,reynders2016,reynders2021}, D\"{o}hler \cite{dohler2013a,dohler2013,mellinger2016} and Gr\'{e}s \cite{gres2022,gres2022a}.
 
Given the development of new UQ techniques for OMA, researchers are also beginning to explore the use of uncertainty to address key research problems in OMA, such as automated modal identification \cite{pereira2020,priou2022} and model order selection \cite{gres2023}. Despite many advancements in UQ for SSI, the authors are currently unaware of a Bayesian formulation of SSI in the literature, where Bayesian posterior distributions over the modal properties are recovered. This article seeks to address this shortcoming.


\subsection{Bayesian uncertainty quantification for OMA}

This section introduces the current landscape of Bayesian approaches to OMA, highlighting associated merits and limitations of existing methods. The major waypoints here include the popular frequency-domain method BAYOMA \cite{au2013} and more recent alternative approaches. These include Gibbs sampling and variational schemes in the time domain using subspace methods, and a Gibbs sampling approach based in the frequency domain.

Bayesian approaches to output-only system identification have been developed since the late 1990s but the topic continues to be of significant interest to the engineering community. Bayesian approaches to OMA are more recent, with the earliest works appearing in late 2000s. The most cited Bayesian approach to OMA follows from the work of Au, whose development of a fast Bayesian Fast Fourier transform algorithm (fast-BFFTA) for modal identification \cite{au2012, au2012a} removed the computational limitations of the original Bayesian fast Fourier transform algorithm \cite{yuen2002,yuen2003} and subsequently led to a body of work known as Bayesian OMA or `BAYOMA' \cite{au2012b, au2013,au2016,au2017,au2018}. BAYOMA can be perceived as the coupling of the fast-BFFTA with a frequency-domain modal analysis technique, such as FDD \cite{brincker2001}. This coupling can be used to recover modal estimates in the form of a most-probable value and a representation of the uncertainty known as the coefficient of variation. These estimates originate from Laplacian approximations to the uncertainty which are used to obtain Gaussian distributions. Since the initial definition of BAYOMA, several extensions have emerged in the literature providing a more general framework to use this methodology, including work on the identification of close-modes \cite{zhu2018,zhu2018a,zhu2020,zhu2021}.

BAYOMA appears to divide some of the research community in its Bayesian definition. In the construction of the algorithm, a flat improper prior is used. Under this assumption, some argue (see \cite{reynders2016}) the problem fundamentally reduces to a maximum likelihood approach, removing the benefits of a Bayesian formulation. Nevertheless, others adopt BAYOMA as their preferred method of choice \cite{mao2023,brownjohn2018,brownjohn2019,yan2015}. Undoubtedly, BAYOMA provides a computationally effective way to perform one form of UQ on structures of interest but is limited in its approximation of the posterior and improper priors.

In search of Bayesian approaches that empower the user to use proper (and informative) prior information, a few alternative Bayesian OMA approaches have been published in recent years. Li and Der Kiureghian \cite{li2017} proposed a variational methodology for OMA and stochastic state space models, where the joint distributions over the state-transition and observation matrices, and the process noise and measurement error, are calculated analytically. These analytical solutions are then coupled with a first-order Taylor series expansion to recover Gaussian approximations to the distributions over the modal properties. This is necessary given the intractability of the posteriors over the modal properties because of the eigenvalue decomposition involved in their recovery \cite{li2017}.

Another Bayesian approach was also presented by Li et.al \cite{li2018a}. The contribution focused on a new form of Bayesian OMA for civil structures under small or moderate seismic excitation. A probabilistic model is first defined, taking advantage of the state space representation of the equations of motion, with some unmeasured base motion included in the model as a stochastic process. Proper and broadly uninformative priors are then introduced into the model, with the Bayesian inference problem solved using a Gibbs sampling procedure to obtain approximate distributions over the modal properties. The two aforementioned recent methodologies differ from those described later in this article, in that they are constructed directly using stochastic state space models and not the SSI-Cov algorithm. 

Lastly, a fast-collapsed Gibbs sampling approach to OMA was introduced by Dollon et.al. \cite{dollon2022}, as a new frequency domain approach. This alternative proposal performs inference on the modal properties using established sampling techniques and the FFT of a system with well-separated modes. The analysis results in sampled posterior distributions that are characterised by a mean and covariance, rather than a most-probable value and coefficient of variation like BAYOMA. 

It is clear that there is considerable interest in developing Bayesian methodologies for performing UQ in OMA. Specifically, methods that can incorporate proper and informative prior structures, and that recover posteriors over the modal properties as distributional estimates. As such, the authors of this paper seek to provide a new approach to Bayesian OMA that meets these requirements.

\subsection{Contribution}\label{sec:Contribution}

This article presents a new perspective on Bayesian OMA, viewed through the lens of the time-domain modal analysis method covariance-driven SSI (SSI-Cov). This new approach is implemented using latent projections in a Bayesian framework, where canonical correlation analysis (CCA) \cite{hotelling1936} ---  a fundamental statistical tool at the core of the canonical-variate weighted SSI-Cov algorithm --- is replaced with a hierarchical Bayesian formulation. This now Bayesian SSI algorithm is capable of recovering posterior distributions over the observability and controllability matrices and, by extension, posteriors over the modal properties.

Two inference schemes for the proposed Bayesian formulation of SSI are provided. The first approach is a Gibbs sampling scheme that recovers an empirical representation of the posterior, which asymptotically trends towards the true posterior with an increasing number of samples. The second approach is a variational scheme that provides a computationally efficient alternative to Gibbs sampling but instead recovers surrogate posteriors that approximate the true posterior. 

To assess the identification and UQ performance of the proposed Bayesian formulation, two case studies are presented. The first is a benchmark study, using data simulated from a linear multi degree-of-freedom system. This initial study is used to compare the posteriors recovered using the variational and Gibbs sampling schemes. It is shown that both schemes target and recover the same posterior distributions over the modal properties, with means closely aligned to the conventional SSI result. The second study analyses data from an in-service structure, specifically the Z24 bridge. The Z24 is chosen as a suitable dataset given its familiarity and frequent use in the fields of OMA and SHM. This study is designed to demonstrate the identification and UQ performance on experimentally obtained data. The results of this study are presented at single model orders, and multiple model orders using a stabilisation diagram. In both cases, it is observed that posterior distributions with mean values coinciding with the apparent natural frequencies exhibit much lower variance than posteriors situated further away from the natural frequencies. 

The remainder of this article is structured as follows. In \textbf{Section \ref{sec: Theory}} the underlying theory of CCA is explored (Section \ref{sec: cca}) and relevant prerequisite topics are outlined, including probabilistic CCA (Section \ref{sec: pcca}) and probabilistic SSI (Section \ref{sec: ProbSSI}). \textbf{Section \ref{sec: Bayesian SSI}} introduces the proposed Bayesian SSI algorithm. The theory of Bayesian CCA is first described, before the Bayesian interpretation of SSI is established. Methods for the recovery of the posterior distributions over the modal properties are then discussed, considering a Gibbs sampling scheme and a variational Bayes approach. A concise stepwise summary of the proposed Bayesian SSI methodology is provided in Section \ref{sec: Summary}. \textbf{Section \ref{sec: case study}} presents the results of a numerical case study. Simulated data were generated and then analysed using the Bayesian SSI algorithm. Following a description of the priors, results from both inference schemes are shown and compared, with an exploration of the influence of data length on the recovered variance. In \textbf{Section \ref{sec: application}}, results from the application of the Bayesian SSI algorithm to data from the Z24 bridge are presented. This includes results at single model orders, and in the form of a stabilisation diagram. Finally, conclusions and future work are discussed in \textbf{Section \ref{sec:Conclusion}}.


\section{Theory} \label{sec: Theory}

\subsection{Canonical correlation analysis}\label{sec: cca}

It is beneficial to briefly review the theory of canonical correlations. CCA is a well established statistical tool developed by Hotelling \cite{hotelling1936} which forms the mathematical basis of canonical variate-weighted SSI-Cov \cite{katayama2005}. The theory of canonical variate-weighted SSI-Cov is shown in \ref{A: SSI-Cov deriv}. For the purpose of this paper, the acronym SSI-Cov will refer to the canonical variate-weighted form. The task of CCA is to analyse the mutual dependency between two multivariate sets of data, which can be evaluated by finding an appropriate set of orthogonal basis vectors, $\vec{a}$ and $\vec{b}$, such that the correlations of the projected variables $\vec{a}^{\trans}\vec{x}$ and $\vec{b}^{\trans}\vec{y}$, are maximally correlated. One then seeks several pairs of vectors that meet the above condition, subject to the constraint that the pairs of transformed variables are uncorrelated from one another. This can be achieved by the following maximisation 

\begin{equation}
	({\vec{a}^{\prime}}, {\vec{b}^{\prime}}) \ = \ \argmax{{\vec{a}},{\vec{b}}} \ \ \mathrm{corr}(\vec{a}^{\trans}\boldsymbol{x},\vec{b}^{\trans}\boldsymbol{y}) \ = \ \argmax{{\vec{a}},{\vec{b}}} \ \dfrac{\vec{a}^{\trans}\mat{\Sigma}_{xy}\vec{b}}{\sqrt{\vec{a}^{\trans}\mat{\Sigma}_{xx}\vec{a}\vec{b}^{\trans}\mat{\Sigma}_{yy}\vec{b}^{\trans}}}
\end{equation}

where $\vec{\Sigma}_{xy}$ is defined as the cross covariance between $\boldsymbol{x}$ and $\boldsymbol{y}$, with $\vec{\Sigma}_{xy} = \vec{\Sigma}_{yx}^{\trans}$, and $\vec{\Sigma}_{xx}$ and $\vec{\Sigma}_{yy}$ are the auto covariances. This maximisation can be computed by solving the following generalised eigenvalue problem

\begin{equation}
    \begin{pmatrix}
        0 & \vec{\Sigma}_{xy} \\
        \vec{\Sigma}_{yx} & 0
    \end{pmatrix} 
    \begin{pmatrix}
        \vec{a}\\
        \vec{b} 
    \end{pmatrix} = \lambda
    \begin{pmatrix}
        \vec{\Sigma}_{xx} & 0 \\
        0 & \vec{\Sigma}_{yy}
    \end{pmatrix}
    \begin{pmatrix}
        \vec{a} \\
        \vec{b} 
    \end{pmatrix}
\end{equation}

where $\vec{a}$ and $\vec{b}$ are the eigenvectors and $\lambda$ is the eigenvalue or `canonical correlation'. In practice, the set of eigenvalues and eigenvectors are computed using the following singular value decomposition (SVD)

\begin{equation}
	\vec{\Sigma}_{xx}^{-\frac{1}{2}}\vec{\Sigma}_{xy}\vec{\Sigma}_{yy}^{-\frac{\trans}{2}} = \mat{V}_1 \mat{\Lambda} \mat{V}_2^{\trans}.
	\label{eq: SVD CCA}
\end{equation}

where $\vec{V}_1$ and $\vec{V}_2$ are left and right singular vectors respectively and $\vec{\Lambda}$ is the matrix of singular values.


\subsection{Probabilistic CCA}\label{sec: pcca}

\begin{figure}[H]
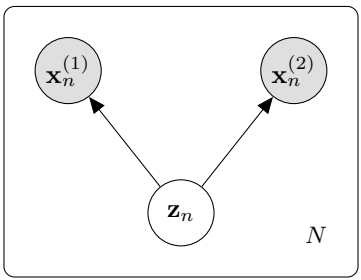

	\centering
	  \tikz{
		\tikzstyle{latent} = [circle,fill=white,draw=black,inner sep=1pt,
		minimum size=25pt, font=\fontsize{10}{10}\selectfont, node distance=1]
		 \node[latent] (z) {$\vec{z}_n$};%
		  \node[obs,above=of z, xshift=-1.5cm] (x1) {$\vec{x}_n^{(1)}$}; %
		  \node[obs,above=of z, xshift=1.5cm] (x2) {$\vec{x}_n^{(2)}$}; %
		\tikzstyle{plate caption} = [caption, node distance=0, inner sep=0pt,
			 below left=-8pt and 0pt of #1.south east]
		  \plate [inner sep=0.4cm] {plate1} {(z)(x1)(x2)} {$N$}; %
		  \edge {z} {x1,x2}  
		  } 
	 \caption{Graphical model for the probabilistic, latent variable interpretation of CCA (PCCA)}
	 \label{fig:Graph PCCA}
\end{figure}

In 2005, Bach and Jordan \cite{bach2005} presented an alternative probabilistic formulation of CCA (PCCA), constructed using latent projections. At the heart of PCCA lies two observed variables $\vec{x}_n^{(1)} \elmtreal{D_1}$ and $\vec{x}_n^{(2)} \elmtreal{D_2}$ which are believed to be conditioned on a lower-dimensional latent space, described by the variable $\vec{z}_n \elmtreal{d}$. It was proposed that the two sets of observed variables can be modelled by an independent linear mapping $\wght{(m)} \elmtreal{D_m \times d}$ of the shared latent variable to the relevant data spaces, plus some mean offset $\vec{\mu}^{(m)} \elmtreal{D_m}$ and with covariance $\vec{\Sigma}^{(m)} \elmtreal{D_m \times D_m}$, for $m = 1, 2$. This model is illustrated as the directed probabilistic graphical model \cite{koller2009} in Figure \ref{fig:Graph PCCA} and described mathematically by Equations (\ref{eq: distr latent z}) - (\ref{eq: distr x}).

\begin{align}
    \vec{z}_n \ \ &\sim \ \ \mathcal{N}(0,\ident)  \label{eq: distr latent z}\\
    \vec{x}^{(m)}_n | \vec{z}_n \ \ &\sim \ \ \mathcal{N}(\vec{\mathrm{W}}^{(m)}\vec{z}_n + \vec{\mu}^{(m)},\vec{\Sigma}^{(m)}) \label{eq: distr x(m)}\\
    \vec{x}_n | \vec{z}_n \ \ &\sim \ \ \mathcal{N}(\vec{\mathrm{W}}\vec{z}_n + \vec{\mu},\vec{\Sigma}) \label{eq: distr x}
\end{align}

where $\gaussdistr{\cdot}{\cdot}$ corresponds to a Gaussian distribution, $\wght{} = [\wght{(1)};\wght{(2)}]$, $\mean{} = [\mean{(1)};\mean{(2)}]$ and $\covar{}$ is a block-diagonal covariance matrix with $\covar{(1)}$ and $\covar{(2)}$ along the diagonal. The isotropic noise model of the latent space enforces the necessary independence of the variables whilst imposing a maximum correlation condition.

Using this model, the maximum likelihood estimates (MLE) of the model parameters can be obtained. In the case of the linear transformations $\vec{W}$, the MLE solution conveniently relates to the components of the SVD (see Equation (\ref{eq: SVD CCA})), such that

\begin{align}
	\wght{(1)} \ \ &= \ \ \mat{\Sigma}^{1/2}_{11}\mat{V}_1\mat{\Lambda}^{1/2}\\
	\wght{(2)} \ \ &= \ \ \mat{\Sigma}^{1/2}_{22}\mat{V}_2\mat{\Lambda}^{1/2}
\end{align}

\subsection{Probabilistic SSI} \label{sec: ProbSSI}

In previous work by the authors \cite{oconnell2024}, it was shown how SSI-Cov could be reimagined as problem in probabilistic inference, using the theory of probabilistic projections (i.e. PCCA) to replace classical CCA in the traditional SSI-Cov algorithm. When the two datasets, $\vec{X}^{(1)}$ and $\vec{X}^{(2)}$, are the Hankel matrices of the future $\vec{Y}_f$ and past $\vec{Y}_p$ respectively, the MLE of the weights $\wght{(m)}$ are equivalent to the extended observability matrix and controllability matrix transposed, with some arbitrary rotational ambiguity $\mat{r}$, such that

\begin{align}
	\wghtmle{(1)} \ \ &= \ \ \mat{\Sigma}^{1/2}_{ff}\mat{V}_1\mat{\Lambda}^{1/2}\mat{R} \ \ = \ \ \mathcal{O} \label{eq:weight1 prob ssi}\\
	\wghtmle{(2)} \ \ &= \ \ \mat{\Sigma}^{1/2}_{pp}\mat{V}_2\mat{\Lambda}^{1/2}\mat{R} \ \  = \ \ \mathcal{C}^{\trans} \label{eq:weight2 prob ssi}
\end{align}

The assimilation of PCCA and SSI in this way is an important revelation. This now probabilistic SSI algorithm allows for familiar hierarchical modelling techniques and permits the use of an arbitrary prior structure, similar to other probabilistic models. The potential of this model was demonstrated by the authors in \cite{oconnell2024}, where a statistically robust noise model was employed to address the problem of misidentification in OMA, brought on by atypical observations (outliers) in measured responses.

\section{Bayesian SSI} \label{sec: Bayesian SSI}

Given the potential to admit an arbitrary prior structure, naturally a Bayesian formulation of SSI is possible. It is perhaps worth considering what a Bayesian SSI algorithm might look like. The most desirable output from a Bayesian OMA algorithm would constitute posterior distributions over the modal properties, given suitably chosen priors that originate from reasonable assumptions given our knowledge of the classical SSI algorithm. Conveniently, such a model can be achieved by incorporating a Bayesian approach to CCA. This section presents the details of a Bayesian formulation of SSI.

\begin{figure}[h]
	\centering 
	\includegraphics{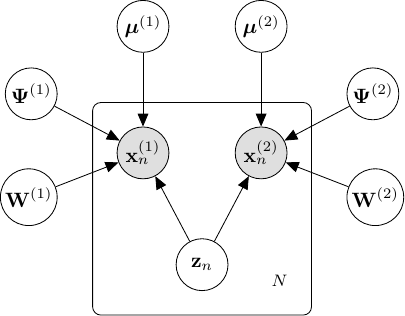}
	\caption{Graphical model for the Bayesian interpretation of CCA}
	\label{fig: bayesian cca graphical model}
\end{figure}

Following the introduction of PCCA, Wang \cite{wang2007} and Klami and Kaski \cite{klami2007} presented the natural hierarchical extension to the PCCA model and introduced Bayesian CCA. This new model incorporated priors over the model parameters $\vec{\theta} = \{\vec{W}, \vec{\Sigma}, \vec{\mu}\}$ and introduced a sparsity inducing prior over the columns of the transformation matrix. The sparsity inducing prior is not included in the Bayesian CCA model described in this paper, nor is it included in later derivations. The inclusion of such a prior for model order selection is of interest however, but will be discussed further in future work. Figure \ref{fig: bayesian cca graphical model} presents the Bayesian CCA graphical model, whilst Equations (\ref{eq: prior wi}) - (\ref{eq: prior Psi}) detail the chosen priors over the model parameters.

\begin{align}
	\vec{w}_i^{(m)} \ \ &\sim \ \ \gaussdistr{\vec{\mu}_{\vec{w}_i}}{\Sigma_{\vec{w}_i}} \label{eq: prior wi}\\
	\vec{W}^{(m)} \ \ &\sim \ \ \prod_{i=1}^{d} p(\vec{w}_i) \\
	\vec{\mu}^{(m)} \ \ &\sim \ \ \gaussdistr{\vec{\mu}_{\mu}}{\Sigma_{\vec{\mu}}} \\ 
	\mat{\Sigma}^{(m)} \ \ &\sim \ \ \iwishdistr{{\mat{K}_0}}{\nu_0} \label{eq: prior Psi}\\
\end{align}

where $\iwishdistr{\vec{K}}{\nu}$ corresponds to an inverse Wishart distribution with scale $\vec{K}$ and $\nu$ degrees of freedom. The precision $\mat{\Psi}$ can also be defined as the Wishart distribution $\mat{\Sigma}^{-1} = \mat{\Psi} \sim \wishdistr{{\mat{K}^{-1}}}{\nu}$. 

Considering the model above, the full joint likelihood can be expressed as
\begin{equation}
	p(\vec{x}, \vec{z}, \vec{W}, \vec{\Sigma}, \vec{\mu}) = p(\vec{x} | \vec{z}, \vec{W}, \vec{\Sigma}, \vec{\mu}) p(\vec{z}) p(\vec{\Sigma}) p(\vec{\mu}) p(\vec{W})
	\label{eq: joint}
\end{equation}

Accounting for the independent columns of W, such that $p(\vec{w}_i)$ describes each column, Equation \ref{eq: joint} becomes
\begin{equation}
	p(\vec{x}, \vec{z}, \vec{W}, \vec{\Sigma}, \vec{\mu}) = p(\vec{x} | \vec{z}, \vec{W}, \vec{\Sigma}, \vec{\mu}) p(\vec{z}) p(\vec{\Sigma}) p(\vec{\mu}) \prod_{i=1}^{d} p(\vec{w}_i) 
\end{equation} 

Given the commonality between the desired weights in PCCA and the desired weights in Bayesian CCA, based on their ability to transform data to the relevant subspace, the posterior distributions over those weights also corresponds to posterior distributions over the observability and controllability matrices in the context of SSI,
\begin{align}
	\mat{W}^{(1)} \ \ &= \ \ \mathcal{O} \label{eq:weight1 bayes ssi}\\
	\mat{W}^{(2)} \ \ &= \ \ \mathcal{C}^{\trans} \label{eq:weight2 bayes ssi}
\end{align}

Unlike the previous definition in Equations (\ref{eq:weight1 prob ssi}) - (\ref{eq:weight2 prob ssi}), here the observability and controllability matrices are now distributional quantities, giving rise to posterior distributions over the modal parameters of interest.

\subsection{Inference schemes}\label{sec: inference}

The key aspect of a Bayesian methodology is the recovery of posterior distributions over the quantities of interest, in this case the modal properties of the structural system. However, as is commonly the case, these distributions are not available in closed form given the form of the model presented above. This intractability of the model arises from inability to compute the integral which defines the normalising constant of the posterior --- the marginal likelihood. Therefore, to solve Bayesian inference problems one must turn to approximate or sampling inferential schemes. 

Two potential solutions commonly arise. The first is to approximate the posterior distribution in a Monte Carlo sense with a number of samples which form an empirical representation of the posterior. These samples can be generated sequentially by means of a Markov Chain, leading to the family of methods known as Markov Chain Monte Carlo (MCMC). Alternatively, the posterior is approximated by a parametric distribution which is defined by the user and described in terms of a set of free parameters. These parameters are then optimised such that the parametric approximation is as close as possible to the true posterior. When a Kullback-Liebler (KL) divergence from the approximate distribution to the true posterior is used, this forms the family of variational inference (VI) methods and the intractable inference problem is transformed into a simpler optimisation problem. 

Within each of these families, a range of approaches exist. In this work, one example from each will be shown and compared. For the MCMC problem, a Gibbs sampling approach is adopted given the direct access to the full conditionals. Gibbs sampling also has a relatively high efficiency in terms of its formulation, whereby every sample proposed is a valid sample from the posterior. For the VI scheme, a coordinate-ascent approach is adopted since the optimisation problem can be solved efficiently in a manner similar to the expectation maximisation (EM) algorithm \cite{murphy2012}. 

Both inferential schemes have their own set of advantages and disadvantages. Gibbs sampling has the very useful property of converging to the true posterior in the limit of increasing number of samples. However, the computational expense of sampling techniques can be significant when considering large and complex joint distributions, making it unsuitable for some tasks. Alternatively, variational methods only provide an approximation to the posterior but demonstrate considerably better computational performance. Hence, both inference approaches are shown in this paper with the results of the inference compared.


\subsubsection{Gibbs sampling}\label{sec: Gibbs Sampling}

Gibbs sampling is a common method in MCMC that can be used to obtain posterior samples of multiple parameters of interest. The resulting distribution of these samples can then be used to approximate the posterior. The premise of Gibbs sampling is that, given a multivariate target distribution, it is much simpler to sample exactly from a conditional distributions rather than marginalising (by integration) over the full joint distribution. For more on Gibbs sampling see e.g. \cite{murphy2012} or \cite{bishop2006}. 

Adopting a Gibbs sampling approach, the parameter sampling equations are derived and are provided in Equations (\ref{eq: gibbs sigma update}) - (\ref{eq: gibbs z update}), with the overall algorithm summarised in Algorithm \ref{al: BayCCA Gibbs algorithm}. For the interested reader, the full derivation of these updates is provided in \ref{A: gibbs}, with the omission of the sparsity inducing prior present in \cite{klami2007} and with the addition of generic mean and covariance priors. 
The required Gibbs sampling updates for the parameters of the model described in Figure \ref{fig: bayesian cca graphical model}, are given below.

\subsubsection*{$\vec{\Sigma}$ update}
\begin{equation}
	\vec{\Sigma}^{(\tau + 1)} \sim \iwishdistr{\vec{K}_0 + \vec{K}}{\nu_0 + N} \label{eq: gibbs sigma update}
\end{equation}
	where 
\begin{equation*}
	\vec{K} = \sum_{n=1}^N (\vec{x}_n - \vec{\mu} - \vec{W}\vec{z}_n ) (\vec{x}_n - \vec{\mu} - \vec{W}\vec{z}_n)^\trans
\end{equation*}
\subsubsection*{$\vec{\mu}$ update}
\begin{equation}
	\vec{\mu}^{(\tau + 1)} \sim \gaussdistr{\hat{\vec{\mu}}_{\mu}}{\hat{\vec{\Sigma}}_{\mu}} \label{eq: gibbs mu update}
\end{equation}
where 
\begin{equation*}
	\hat{\vec{\Sigma}}_{\mu} = (N\vec{\Sigma}^{-1} + \vec{\Sigma}_{\mu}^{-1})^{-1}
\end{equation*}
and
\begin{equation*}
	\hat{\vec{\mu}}_{\mu} = \hat{\vec{\Sigma}}_{\mu}\left(\vec{\Sigma}^{-1} \sum_{n=1}^N (\vec{x}_n - \vec{W}\vec{z}_n) + \vec{\Sigma}_{\mu}^{-1}\vec{\mu}_{\mu}\right)
\end{equation*}
\subsubsection*{$\vec{w}_i$ update}
\begin{equation}
	\vec{w}_i^{(\tau + 1)} \sim \gaussdistr{\hat{\vec{\mu}}_{\vec{w}_i}}{\hat{\vec{\Sigma}}_{\vec{w}_i}} \label{eq: gibbs wi update}
\end{equation}
where
\begin{equation*}
	\hat{\vec{\Sigma}}_{\vec{w}_i} = \left (\sum_{n=1}^N  \vec{z}_{i,n}^2 \vec{\Sigma}^{-1} + \vec{\Sigma}_{\vec{w}_i}^{-1} \right )^{-1}
\end{equation*}
and
\begin{equation*}
	\hat{\vec{\mu}}_{\vec{w}_i} = \hat{\vec{\Sigma}}_{\vec{w}_i}(\vec{\Sigma}^{-1} \sum_{n=1}^N \vec{z}_{i,n}\tilde{\vec{x}}_n + \vec{\Sigma}_{\vec{w}_i}^{-1}\vec{\mu}_{\vec{w}_i})
\end{equation*}
given
\begin{equation*}
	\tilde{\vec{x}}_n  = \vec{x}_n - \vec{\mu} - \vec{w}_{\_i}\vec{z}_{\_i,n}
\end{equation*}
such that $\vec{w}_{\_i}$ corresponds to all columns except the $i$th column of interest and $\vec{z}_{\_i,n}$ corresponds to all rows of $\vec{z}_n$ except the $i$th row. 

\subsubsection*{$\vec{z}$ update}
\begin{equation}
	\vec{z}_n^{(\tau + 1)} \sim \gaussdistr{\hat{\vec{\mu}}_{\vec{z}}}{\hat{\vec{\Sigma}}_{\vec{z}}} \label{eq: gibbs z update}
\end{equation}
	where 
\begin{equation*}
	\hat{\vec{\Sigma}}_{\vec{z}} = (\vec{W}^\trans\vec{\Sigma}^{-1}\vec{W} + \vec{\ident})^{-1}
\end{equation*}
and
\begin{equation*}
	\hat{\vec{\mu}}_{\vec{z}} = \hat{\vec{\Sigma}}_{\vec{z}}\vec{W}^\trans\vec{\Sigma}^{-1}\vec{x}_n
\end{equation*}

\begin{algorithm}[H]
	\caption{Bayesian CCA - Gibbs Sampling (MCMC)}
	\begin{algorithmic}
	\State \textbf{initialise}: $\vec{\Sigma},\vec{\mu},\vec{W},\vec{z}$ using the priors given appropriately chosen hyperparameters
	\For {$\tau=1 \dots, T$}
		\State Sample $\vec{\Sigma}^{(\tau + 1)}$ using Equation (\ref{eq: gibbs sigma update})
		\State Sample $\vec{\mu}^{(\tau + 1)}$ using Equation (\ref{eq: gibbs mu update})
		\For {$i=1 \dots, d$}
			\State Sample $\vec{w}_i^{(\tau + 1)}$ using Equation (\ref{eq: gibbs wi update})
		\EndFor
		\State Sample $\vec{z}_n^{(\tau + 1)}$ using Equation (\ref{eq: gibbs z update})
	\EndFor\\
	\Return Samples of $\vec{\Sigma},\vec{\mu},\vec{W},\vec{z}$
	\end{algorithmic}
	\label{al: BayCCA Gibbs algorithm}
\end{algorithm}

Following application of Algorithm \ref{al: BayCCA Gibbs algorithm} for a desired number of samples, the resulting samples of $\mat{W}^{(1)}$ (i.e. the observability matrix) can be propagated to obtain samples of the state transition matrix and subsequently propagated further onto the modal properties in the usual way for a linear dynamic system \cite{katayama2005}, obtaining sampled posterior distributions. 

Despite the useful property of MCMC techniques -- guaranteed convergence to the true posterior in the limit of the number of samples -- such techniques can be inherently slow and computationally inefficient, especially if the posteriors are highly correlated in Gibbs samplers.

\subsubsection{Variational Bayes}\label{sec:Bayesian CCA Var}

VI is a well-known form of Bayesian inference and common alternative to MCMC, used to approximate intractable posterior distributions with analytical approximations (see e.g. \cite{murphy2012}, \cite{bishop2006}). The general premise of VI is to select a suitable approximation (surrogate) from a tractable family of distributions and try to make the approximation as close to the true (intractable) posterior as possible. This is achieved by minimising the KL divergence $\mathbb{K}\mathbb{L}(q||p)$ from the surrogate distribution $q$ to the true posterior distribution $p$. The minimisation of the KL divergence is equivalent to maximising the evidence lower bound (ELBO), which is equal to the negative KL divergence up to an additive constant. The ELBO can also be defined as the sum of the expected log of the joint distribution, and the entropy of the variational distribution.

Assuming the surrogate posterior is determined by some free parameters, then the problem of VI reduces to a more familiar optimisation problem \cite{murphy2012}. Which, for certain cases, can be solved using coordinate-ascent and conducted efficiently in a manner similar to the expectation maximisation (EM) algorithm \cite{murphy2012}. When latent variables $\vec{z}_n$ and the parameters of the model $\vec{\theta}$ are desired, as is the case in Bayesian CCA, then the method is known as variational Bayes (VB). 

An important step in VB is defining the form of the surrogate posterior. Here the mean-field approximation \cite{Opper2001} is chosen, where the posterior can be fully factorised in the form $q(\mathrm{z}) = q(\theta) \prod_{j=1}^{J}q_j(\mathrm{z}_j)$. Adopting a mean field approximation, the surrogate posterior for Bayesian CCA takes the following factorised form

\begin{equation}
	q(\vec{z}, \vec{\mu}, \vec{\Psi}, \vec{W}) = q(\vec{z})\prod_{i=1}^{d}q(\vec{w}_i)q(\vec{\mu})q(\vec{\Psi}) \label{eq: meanfield}
\end{equation}
where 
\begin{align}
    q(\vec{z}_n) \ \ &\sim \ \ \gaussdistr{\vec{z}_n | \breve{\vec{\mu}}_{\vec{z}}}{\breve{\vec{\Sigma}}_{\vec{z}}} \\
    q(\vec{\Psi}) \ \ &\sim \ \ \wishdistr{\vec{\Psi} | \breve{\vec{K}}^{-1}}{\breve{\nu}} \\
    q(\vec{w}_i) \ \ &\sim \ \ \gaussdistr{\vec{w}_i | \breve{\vec{\mu}}_{\vec{w}_i}}{\breve{\vec{\Sigma}}_{\vec{w}_i}} \\
    q(\vec{\mu}) \ \ &\sim \ \ \gaussdistr{\vec{\mu} | \breve{\vec{\mu}}_{\vec{\mu}}}{\breve{\vec{\Sigma}}_{\vec{\mu}}}
\end{align}

Following the VB methodology, the update equations for the model parameters were derived and are given in Equations (\ref{eq: var z update})-(\ref{eq: var mu update}). The complete algorithm is summarised in Algorithm \ref{al: BayCCA VI algorithm}, whilst the full derivation of these updates is also provided in \ref{A: VI}, with the omission of the sparsity inducing prior present in \cite{wang2007} and with generic mean and covariance priors.

The variational update equations for the parameters of the model described in Figure \ref{fig: bayesian cca graphical model}, are given below. In all equations, $\langle \cdot \rangle$ represents the expected value $\expect[]{\cdot}$ of that variable or combination of variables with respect to all the other parameters.

\subsubsection*{$\vec{z}_n$ update}
\begin{equation}
	q^{\star}(\vec{z}_n) \sim \gaussdistr{\breve{\vec{\mu}}_{\vec{z}}}{\breve{\vec{\Sigma}}_{\vec{z}}} \label{eq: var z update}
\end{equation}
where 
\begin{equation*}
	\breve{\vec{\Sigma}}_{\vec{z}} = \left(\langle \vec{W}^\trans \vec{\Psi}\vec{W}\rangle  + \vec{\ident} \right)^{-1}
\end{equation*}
and 
\begin{equation*}
	\breve{\vec{\mu}}_{\vec{z}} = \breve{\vec{\Sigma}}_{\vec{z}} \langle \vec{W}\rangle ^\trans\langle \vec{\Psi}\rangle (\vec{x}_n - \langle \vec{\mu}\rangle )
\end{equation*}

\subsubsection*{$\vec{w}_i$ update}
\begin{equation}
	q^{\star}(\vec{w}_i) \sim \gaussdistr{\breve{\vec{\mu}}_{\vec{w}_i}}{\breve{\vec{\Sigma}}_{\vec{w}_i}} \label{eq: var wi update}
\end{equation}
where 
\begin{equation*}
	\breve{\vec{\Sigma}}_{\vec{w}_i} = \left(\sum^{N}_{n=1} \langle \vec{z}_{i,n}\vec{z}_{i,n} \rangle \langle\vec{\Psi}\rangle + \vec{\Sigma}_{\vec{w}_i}^{-1}\right)^{-1}
\end{equation*}
and 
\begin{equation*}
	\breve{\vec{\mu}}_{\vec{w}_i} = \breve{\vec{\Sigma}}_{\vec{w}_i}\left(\langle\vec{\Psi}\rangle \sum_{n=1}^N \tilde{\vec{x}}_n\langle\vec{z}_{n,i}^{\trans}\rangle + \vec{\Sigma}_{\vec{w}_i}^{-1}\vec{\mu}_{\vec{w}_i}\right)
\end{equation*}
given
\begin{equation*}
	\tilde{\vec{x}}_n  = \vec{x}_n - \langle\vec{\mu}\rangle - \langle\vec{w}_{\_i}\rangle\langle\vec{z}_{\_i,n}\rangle
\end{equation*}
such that $\vec{w}_{\_i}$ corresponds to all columns except the $i$th column of interest and $\vec{z}_{\_i,n}$ corresponds to all rows of $\vec{z}_n$ except the $i$th row. 

\subsubsection*{$\vec{\Psi}$ update}
\begin{equation}
	q^{\star}(\vec{\Psi}) \sim \wishdistr{\breve{\vec{K}}^{-1}}{\breve{\nu}} \label{eq: var Psi update}
\end{equation}
where 
\begin{equation*}
	\breve{\vec{K}} = \vec{K}_0 +  \sum_{n=1}^N \left\langle (\vec{x}_n - \vec{\mu} - \vec{W}\vec{z}_n ) (\vec{x}_n - \vec{\mu} - \vec{W}\vec{z}_n)^\trans \right\rangle 
\end{equation*} 
and 
\begin{equation*}
	\breve{\nu} = \nu_0 + N
\end{equation*}

\subsubsection*{$\vec{\mu}$ update}
\begin{equation}
	q^{\star}(\vec{\mu}) \sim \gaussdistr{\breve{\vec{\mu}}_{\vec{\mu}}}{\breve{\vec{\Sigma}}_{\vec{\mu}}} \label{eq: var mu update}
\end{equation}
where 
\begin{equation*}
	\breve{\vec{\Sigma}}_{\vec{\mu}} = (N\langle\vec{\Psi}\rangle + \vec{\Sigma}_{\mu}^{-1})^{-1}
\end{equation*}
and 
\begin{equation*}
	\breve{\vec{\mu}}_{\vec{\mu}} = \breve{\vec{\Sigma}}_{\mu}\left(\langle\vec{\Psi}\rangle \sum_{n=1}^N (\vec{x}_n - \langle\vec{W}\rangle\langle\vec{z}_n\rangle) + \vec{\Sigma}_{\mu}^{-1}\vec{\mu}_{\mu}\right)
\end{equation*}

\begin{algorithm}[H]
	\caption{Bayesian CCA - Coordinate Ascent Variational Inference (CAVI)}
	\begin{algorithmic}
	\State \textbf{initialise}: Variational factors $q(\vec{z}, \vec{\mu}, \vec{\Psi}, \vec{W})$
	\While {the $\mathrm{ELBO}$ has not converged}
		\State \textbf{Update local variational parameters}:
		\State Update $q^{\star}(\vec{z}_n)$ using Equation (\ref{eq: var z update})\\

		\State \textbf{Update global variational parameters}:
		\For {$i = 1, \dots, d$}
			\State Update $q^{\star}(\vec{w}_i)$ using Equation (\ref{eq: var wi update})
		\EndFor
		\State Update $q^{\star}(\vec{\Psi})$ using Equation (\ref{eq: var Psi update})
		\State Update $q^{\star}(\vec{\mu})$ using Equation (\ref{eq: var mu update}) \\

		\State Compute $\mathrm{ELBO}(q) = \expect{\ln{p(\vec{x}, \vec{z}, \vec{\theta})}} + \expect{\ln{q(\vec{z},\vec{\theta})}}$
	\EndWhile\\
	\Return Surrogate posterior distributions of $\vec{\Sigma},\vec{\mu},\vec{W},\vec{z}$
	\end{algorithmic}
	\label{al: BayCCA VI algorithm}
\end{algorithm}


\subsection{Overall methodology}\label{sec: Summary}

Having explored two different forms of inference for the recovery of the posteriors in Bayesian CCA, it is perhaps useful to summarise the general procedure in the context of Bayesian SSI. The following summary gives an overview of the steps needed to conduct Bayesian SSI to a desired dataset.

Given response data, $\vec{y} \elmtreal{d_{s}\times N}$, with $d_{s}$ sensors and $N$ datapoints:\\

\textbf{Step 1.} Construct Block Hankel matrices of the future, $\vec{Y}_f \elmtreal{D_m \times N}$, and the past, $\vec{Y}_p \elmtreal{D_m \times N}$,\\ where $D_m = d_s \times d_{\mathrm{Hankel}}$ given that $d_{\mathrm{Hankel}}$ is the chosen number of lags in the Hankel matrix.\\

\textbf{Step 2.} Apply \textbf{Algorithm 1} or \textbf{Algorithm 2} with the inputs $\vec{x}_1 = \vec{Y}_f$ and $\vec{x}_2 = \vec{Y}_p$, given appropriately chosen priors and convergence criteria\footnote{The convergence criteria, or desired number of samples, in both algorithms must be selected. The choice and specification of these in optimisation tasks constitutes its own research field and are therefore not explored in depth.} to obtain the posteriors over the model parameters.\\

\textbf{Step 3.} If \textbf{Algorithm 1} has been applied, remove a proportion of the initial samples (as burn in) and continue to Step 4. If \textbf{Algorithm 2} has been applied, draw Monte Carlo samples from the posterior distributions over the columns of the weight matrix and reconstruct the full weight matrix to provide a sample of the observability, i.e $\mat{W}_1$.\\

\textbf{Step 4.} Using the samples of the observability matrix, calculate the state transition matrix\footnote{There are multiple approaches used to do this step. The authors adopt the balanced truncation algorithm, see Chapter 8 in Katayama \cite{katayama2005}} $\vec{A}$ and the modal properties in the usual way for state space models (e.g. Section 3.3 in \cite{reynders2008}), for each sample. Considering all the samples, this provides an approximate posterior distribution over each modal property.

\section{Simulated case study}\label{sec: case study}

\subsection{Benchmarking the proposed algorithm}\label{sec: benchmark}

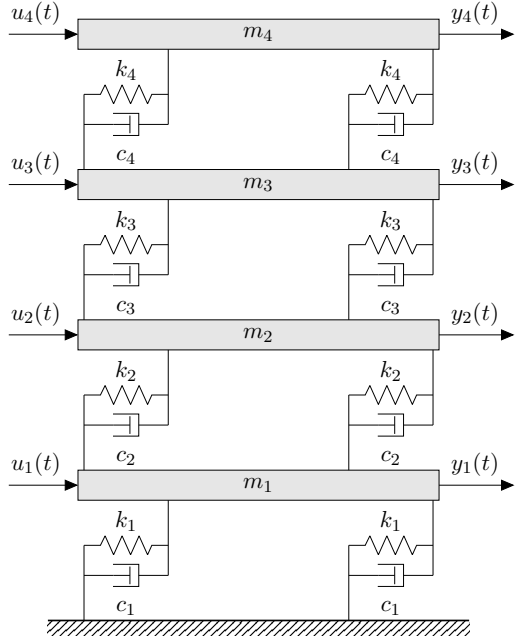
\begin{figure}[H]
	\begin{center}
	\scalebox{0.8}{
		\begin{circuitikz}

		\ctikzset{resistors/scale=0.75}
		\ctikzset{resistors/thickness=1}
		\ctikzset{mechanicals/scale=0.75}
		\ctikzset{mechanicals/thickness=1}
		\ctikzset{mechanicals/thickness=1}
		
		\newcommand*{\newfloor}[1]{
			\pgfmathsetmacro{\floorheight}{#1 * 2.5}
			\pgfmathtruncatemacro{\floornum}{#1 + 1}

			\draw[fill=gray!20] (0.5, 2 + \floorheight) rectangle (6.5, 2.5 + \floorheight);
			\node at (3.5, 2.25 + \floorheight) {$m_{\floornum}$};

			\draw (0.6, 0 + \floorheight) -- (0.6, 1.25 + \floorheight);
			\draw (0.6, 1.25 + \floorheight) to[resistor, l=$k_{\floornum}$] (2, 1.25 + \floorheight);
			\draw(0.6, 0.75 + \floorheight) to[damper, l_=$c_{\floornum}$] (2,0.75  + \floorheight); 
			\draw (2, 0.75  + \floorheight) -- (2, 2  + \floorheight);

			\draw (5, 0  + \floorheight) -- (5, 1.25  + \floorheight);
			\draw (5, 1.25  + \floorheight) to[resistor, l=$k_{\floornum}$] (6.4, 1.25  + \floorheight);
			\draw(5, 0.75  + \floorheight) to[damper, l_=$c_{\floornum}$] (6.4, 0.75  + \floorheight); 
			\draw (6.4, 0.75  + \floorheight) -- (6.4, 2  + \floorheight);

			\draw[->] (6.5, 2.25 + \floorheight) -- (7.75, 2.25 + \floorheight);
			\node at (7.1, 2.6 + \floorheight) {$y_{\floornum}(t)$};

			\draw[->] (-0.65, 2.25 + \floorheight) -- (0.5, 2.25 + \floorheight);
			\node at (-0.2, 2.6 + \floorheight) {$u_{\floornum}(t)$};

		}
		\newcommand*{\ground}{
			\pattern[pattern=north east lines, distance={1pt}] (0,0) rectangle (7,-0.25);
			\draw[thick] (0,-0) -- (7,0);
		}

		\ground
		\newfloor{0}
		\newfloor{1}
		\newfloor{2}
		\newfloor{3}

		\end{circuitikz}
	}
	\end{center}
	\caption{Four-story shear building model used to simulate responses given a white noise input on each floor}
	\label{fig:Simulated MDOF System}
\end{figure}

The performance of the proposed Bayesian SSI algorithm, outlined in Section \ref{sec: Summary}, will be benchmarked here by application to simulated data obtained using a numerical model of the four degree-of-freedom shear frame illustrated in Figure \ref{fig:Simulated MDOF System}. This structure is identical to that described by Reynders in \cite{reynders2021}, although a new simulation is used. Using a mass-spring-damper model to represent the dynamics of the system, the mass $m_j$ assigned to each floor is 2 kg and the stiffness $k_j$ applied to the individual columns (springs) on each floor is 2500 N/m, where $j = 1,2,3,4$. The damping in each column is proportional to the associated stiffness, such that $c_j = k_j / 1000$. Horizontal forces are assumed to apply to each floor as the inputs $u_j(t)$, whilst the horizontal acceleration of each floor are the corresponding outputs $y_j(t)$. 

The equations of motion of the shear frame in continuous time can be defined as the ordinary differential equation
\begin{equation}
	[\vec{M}]\ddot{\vec{x}} + [\vec{C}]\dot{\vec{x}} + [\vec{K}]\vec{x} = \vec{f}
	\label{eq: eq of motion}
\end{equation}

In OMA, the input is unknown and often assumed to be Gaussian white noise. Therefore, the forcing $\vec{f}$ can be defined as a white noise process in continuous time acting on each floor. This results in a linear stochastic differential equation which is then discretised exactly using Van Loan discretisation \cite{sarkka2019a}. The forcing white noise process is chosen with a spectral density of $5\times10^{-5}$ $(\mathrm{m}$ $\mathrm{s}^{-2})^2 \mathrm{Hz}^{-1}$ whilst some Gaussian measurement noise is added to each channel output $y_j(t)$, sampling from a zero-mean normal distribution with a standard deviation of 0.05 $\mathrm{m}$ $\mathrm{s}^{-2}$. Data are then simulated using a sampling frequency of 50 Hz with length $N = 2^{16}$. 

Based on the model described in Figure \ref{fig:Simulated MDOF System}, the matrix coefficients of Equation \ref{eq: eq of motion} are

\begin{eqnarray*}
		\mat{M} = \begin{bmatrix} m_1 & 0 & 0 & 0\\ 0 & m_2 & 0 & 0 \\ 0 & 0 & m_3 & 0 \\ 0 & 0 & 0 & m_4 \end{bmatrix}\ , \ \ 
		\mat{K} = 2 \begin{bmatrix} k_1 + k_2 & -k_2 &0 & 0\\ -k_2 & k_2 + k_3 & -k_3 & 0 \\ 0 & -k_3 & k_3 + k_4 & -k_4 \\ 0 & 0 & -k_4 & k_4 \end{bmatrix} \ , \ \ 
		\mat{C} = \mat{K} / 1000
		\label{eq: MDOF}
\end{eqnarray*}

\subsubsection{Priors}\label{sec: numsim - prior}

\begin{figure}[H]
	\centering 
	\includegraphics[width=\textwidth]{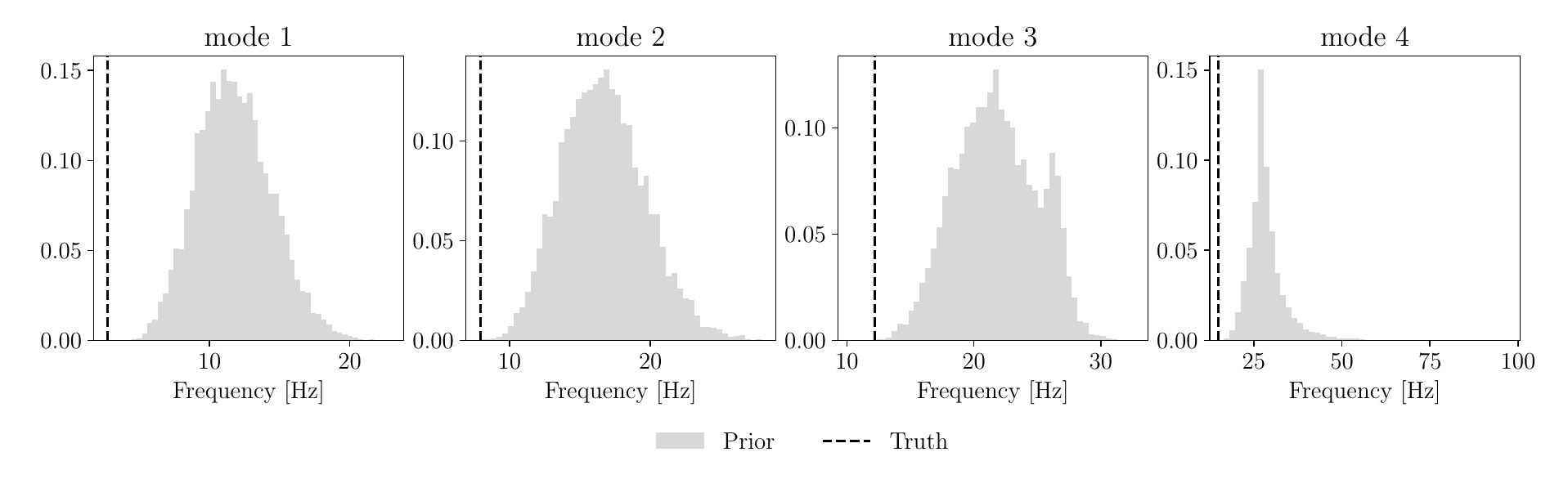}
	\footnotesize (a)
	\includegraphics[width=\textwidth]{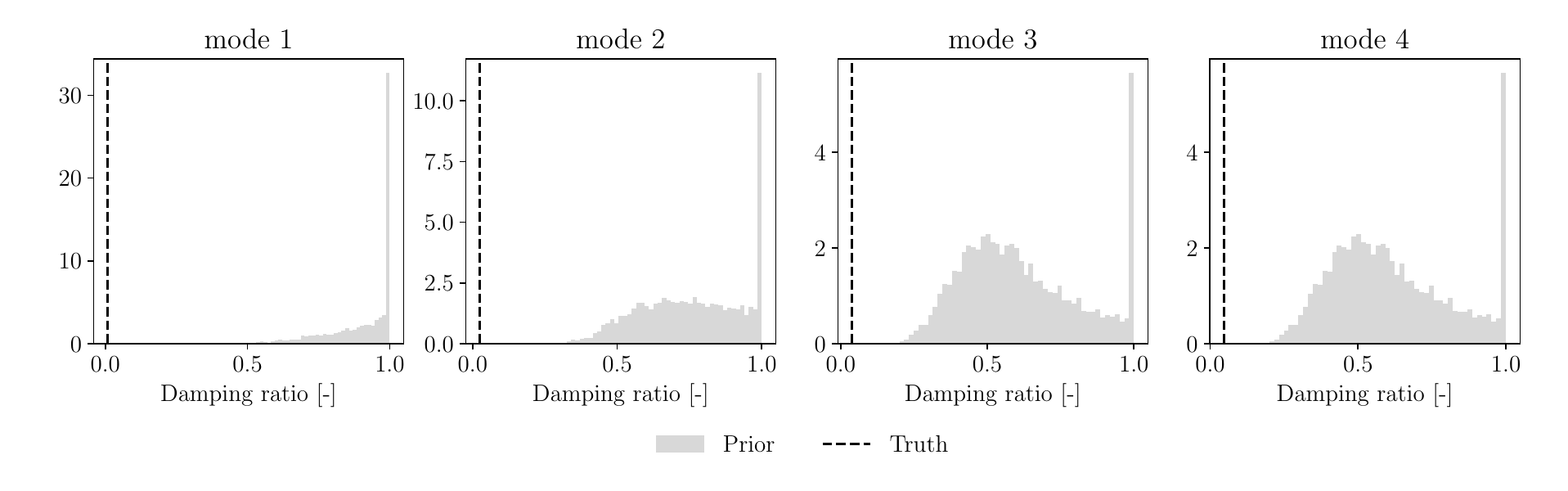}
	(b)
	\includegraphics[width=\textwidth]{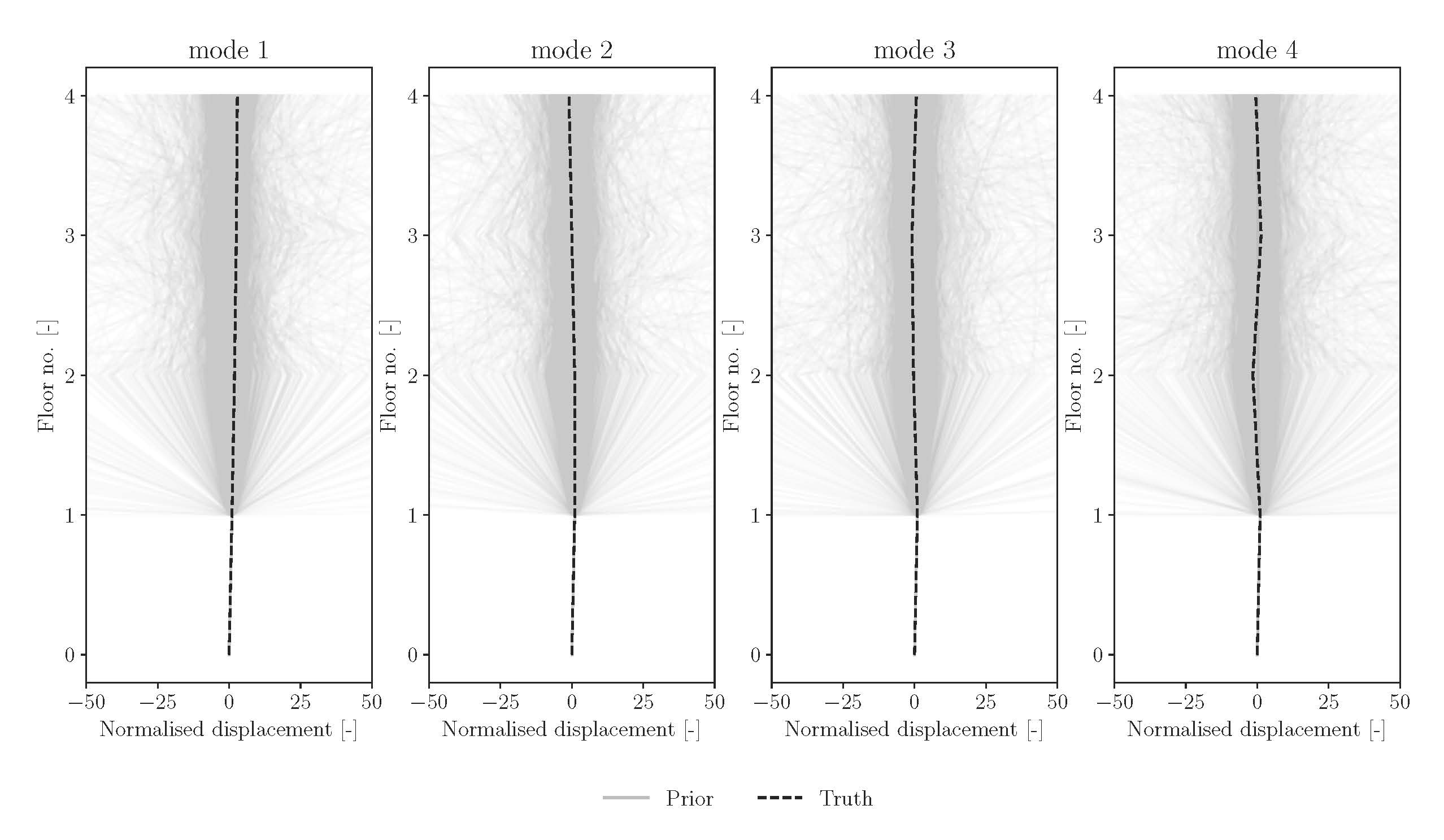}
	(c)
	\caption{Prior distributions over the natural frequencies (a), damping ratios (b) and mode shapes (c), obtained using propagated Monte Carlo samples of the priors as defined by Equations \ref{eq: prior wi} - \ref{eq: prior Psi}.}
	\label{fig: mdof priors}
\end{figure}

Before carrying out any inference, attention must first be paid towards establishing sensible priors for the model. Given the general model presented in Section \ref{sec: benchmark}, the following prior structure was chosen
\begin{align}
	\vec{w}_i^{(m)} \ \ &\sim \ \ \gaussdistr{\vec{0}}{\sigma_{\vec{w}_i}\ident} \label{eq: Prior wi}\\
	\vec{\mu}^{(m)} \ \ &\sim \ \ \gaussdistr{\vec{0}}{\sigma_{\vec{\mu}}\ident}\\ 
	\mat{\Sigma}^{(m)} \ \ &\sim \ \ \iwishdistr{{\mat{K}_0}}{\nu_0}
\end{align}

where the independent columns $\vec{w}_{i}$ share the same prior definition. The remaining hyperparameters, $\vec{\sigma}_{\mu} = \sigma_{\vec{w}_i} = 1$, $\vec{K}_0 = 1 \times 10^{2}$, $\nu_0 = D + 2$, were chosen to be weakly informative and proper, to provide sufficient flexibility to the model. Samples from the chosen prior over the weight matrices were then used to generate samples of the observability matrix and subsequently propagated through an eigenvalue decomposition to obtain estimates for the posterior distributions over the modal properties. The prior distributions over the modal properties are presented in Figure \ref{fig: mdof priors}. 



\newpage

It is useful to briefly discuss the general shape of the priors on the modal parameters. Despite Gaussian assumptions in the model, specifically on the weight prior, the propagated uncertainty results in distributions that are a mix of what appear to be non-Gaussian and even multimodal in nature. The authors believe this is likely caused by the size of the prior variance coupled with the non-linear transformation in the eigenvalue decomposition, which is most evident when considering larger variances. 

\subsubsection{Posteriors}\label{sec: posteriors}

\begin{figure}[H]
	\centering 
	\includegraphics[width=\textwidth]{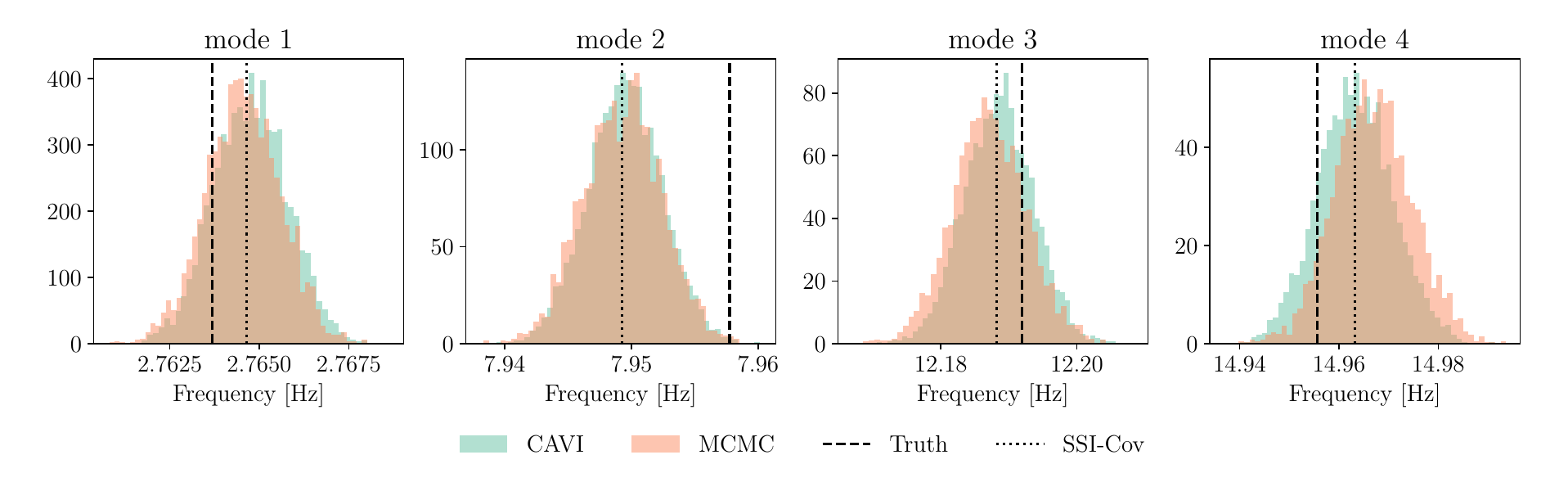}
	\footnotesize (a)
	\includegraphics[width=\textwidth]{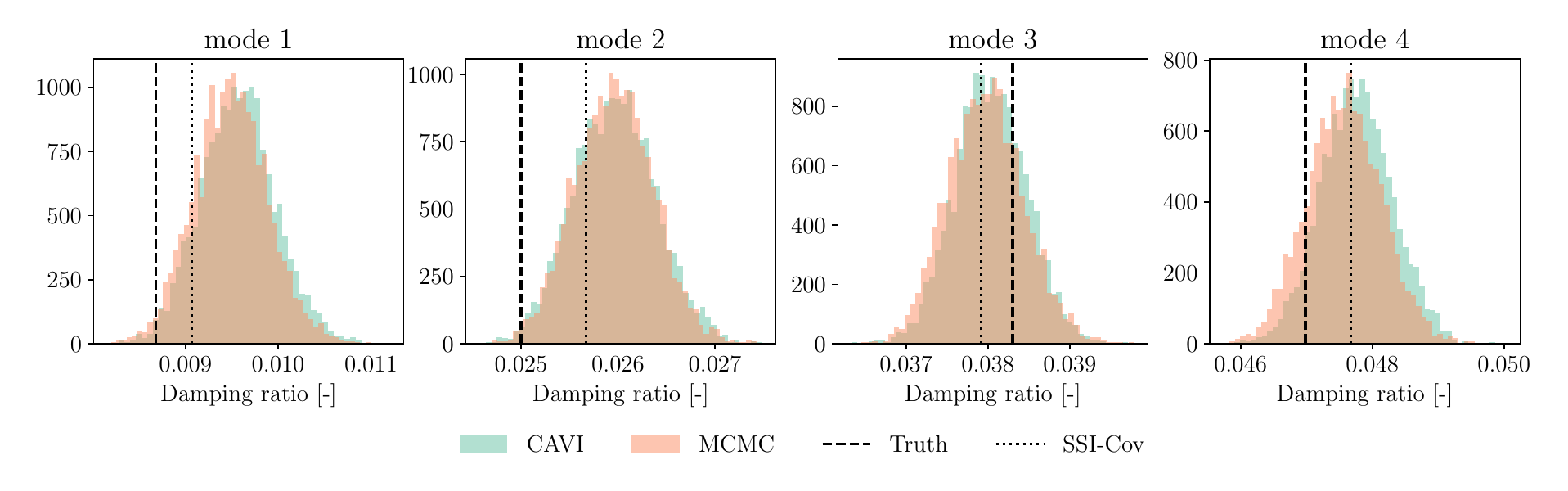}
	(b)
	\includegraphics[width=\textwidth]{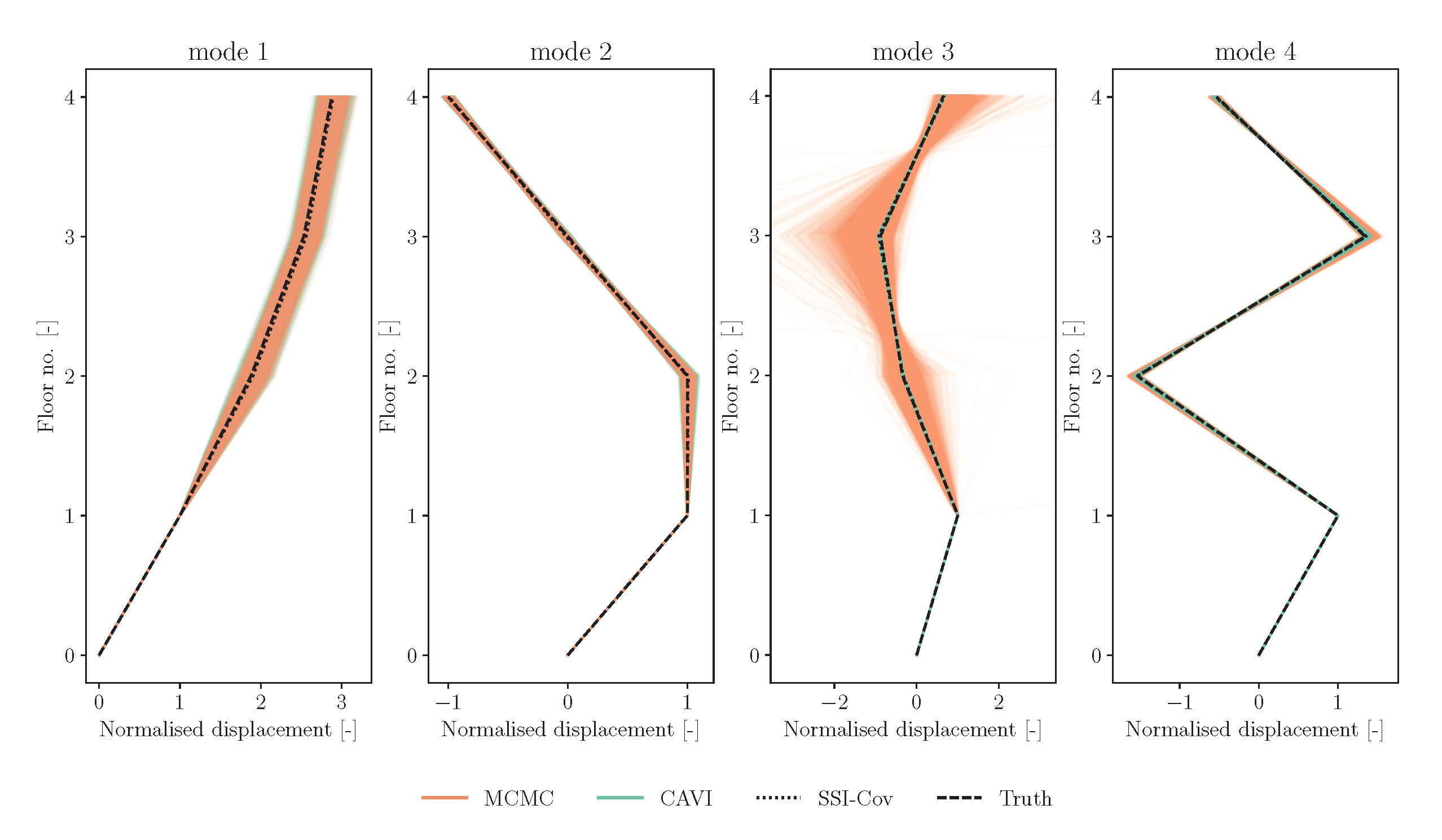}
	(c)
	\caption{Identified posterior distributions over the recovered natural frequencies (a), damping ratios (b) and mode shapes (c), obtained using the Gibbs sampling and VB implementations of Bayesian SSI to response data simulated from a 4DOF linear dynamic system (see Figure \ref{fig:Simulated MDOF System}).}
	\label{fig: mdof posteriors}
\end{figure}

Assuming the correct number of modes (4 modes -- corresponding to a state space dimension of 8), the Gibbs sampling implementation of Bayesian SSI (Algorithm \ref{al: BayCCA Gibbs algorithm}) was used to recover posterior distributions over the modal properties of the simulated structure. 5000 samples were drawn, with the first 20\% removed as `burn in' to remove transients in the Markov chain \cite{murphy2012}. The posteriors over the modal properties are shown in Figure \ref{fig: mdof posteriors}. Similar to the Gibbs sampling case, assuming the correct number of modes, data were analysed using the VI implementation of Bayesian SSI (Algorithm \ref{al: BayCCA VI algorithm}). After the recovery of the closed form posterior over the observability, 4000 samples were drawn and propagated. The resulting posteriors over the modal properties are also presented in Figure \ref{fig: mdof posteriors}.

On interpreting Figure \ref{fig: mdof posteriors}, one can conclude that the surrogate posteriors over the modal parameters obtained using VB closely align with the true posteriors obtained through Gibbs sampling. This closeness is evident from the aligned expected values and the shape of the posteriors, which correspond well. Furthermore, the posterior estimates of both schemes converge toward the SSI-Cov result. This convergence is largely expected as, given the use of weakly informative priors and enough data, the maximum-a-posteriori (MAP) will be close to the MLE, i.e. SSI-Cov. Some minor differences can be seen. The first is in the mean estimates of first and second damping ratios compared to the SSI-Cov result. This misalignment is a consequence of the priors, which cause slight bias in the posterior, more so than the damping ratio estimates for the third and fourth modes. The second difference can be seen in the variance of the third mode shape in Figure \ref{fig: mdof posteriors}c. A non-Gaussian posterior recovered from the Gibbs scheme explains this difference. This phenomenon originates from application of the eigenvalue decomposition to obtain the mode shapes. However, this difference is more likely a result of an underprediction of the variance using VI. Under or overprediction of the variance is a common characteristic of VI schemes which can occur when some conditional distributions are factorised out of the model during its construction \cite{murphy2012}. The omission of these conditionals is a result of the independence assumptions in the surrogate posterior definition. Possible misalignment of the variance is one of several trade-offs that needs to be considered when prioritising the computational efficiency of variational methods over convergence to the true posterior guaranteed by MCMC sampling.

\newpage

\subsection{Influence of data length on variance} \label{sec: varying data length}

\begin{figure}[h!]
	\begin{subfigure}{\textwidth}
		\centering
		\includegraphics[width=\textwidth]{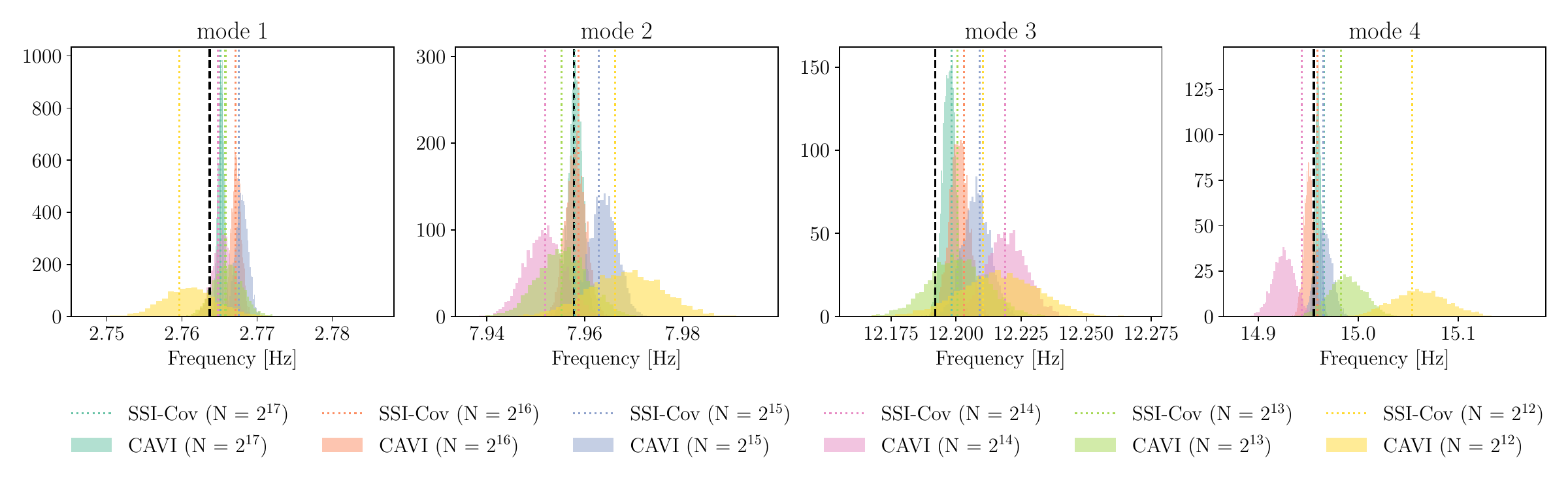}
		\caption{}
		\label{fig:modal variance freq}
	\end{subfigure}
	\begin{subfigure}{\textwidth}
		\centering
		\includegraphics[width=\textwidth]{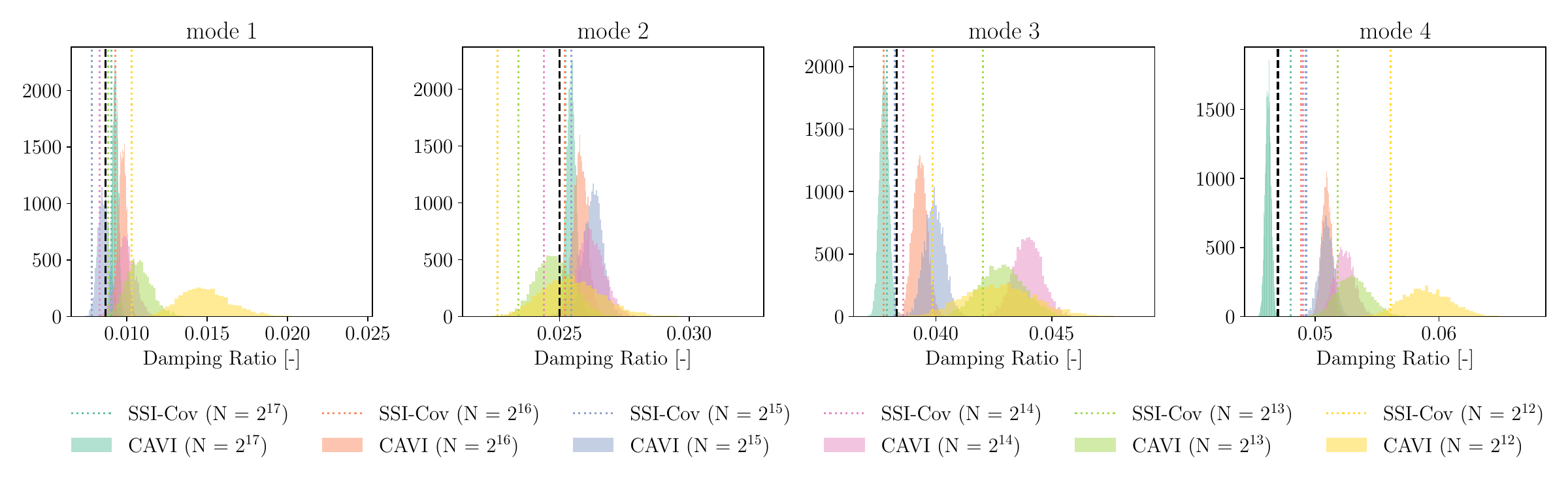}
		\caption{}
		\label{fig:modal variance damp}
	\end{subfigure}
	\caption{Identified posterior distributions over the natural frequencies (a) and damping ratio (b) estimates for each mode, obtained using Bayesian SSI with varying data length N. The true values are represented by \textbf{- -}}
\end{figure}

The influence of data length on the variance of the modal posteriors is also investigated. Using the same system parameters and priors described in the Section \ref{sec: numsim - prior}, the data length was varied from $2^{17} - 2^{12}$ in decreasing powers of 2, with the range selected solely for demonstrative purposes. The results for the natural frequencies and the damping ratios are displayed in Figures \ref{fig:modal variance freq} and \ref{fig:modal variance damp}, respectively. As is perhaps expected, the variance of the posterior estimates of both frequency and damping ratio reduce given increasing amounts of data, meanwhile the means of the distributions also converge towards the true values from the model. The influence of data length and perhaps other factors (lags in the Hankel matrix) on the recovered uncertainty has interesting implications and could be used to reduce the amount of data collection and storage required in OMA, for a desired level of confidence. This poses an interesting question on choosing the right amount of data for a given decision-making task, which will be considered in future work. 

\newpage
\section{Case Study: Z24 bridge}\label{sec: application}

To demonstrate applicability to real world systems, the proposed Bayesian SSI algorithm is used to analyse vibration data obtained from the Z24, a now decommissioned bridge once located between Bern and Zurich, Switzerland. The Z24-Bridge has become a standard benchmark when demonstrating vibration-based damage diagnosis methods, new system identification and OMA techniques, and UQ tasks in structural dynamics, making it a sensible choice for this study. For more information on the testing of the Z24, a full description, initial analysis and subsequent testing, the reader is directed towards (bwk.kuleuven.be/bwm/z24) \cite{maeck2003,mevel2003}. 

This purpose of this study was to evaluate the behaviour of the new Bayesian SSI algorithm when confronted with measured data. Analysis is only conducted using the variational approach, since the high computation time of Gibbs sampler was deemed impractical. Results at two different single model orders are shown, followed by a stabilisation diagram to assess the convergence of the modal properties at different model orders.

Data, corresponding to acceleration measurements obtained on the 24th December 1997, were used for processing. This dataset comes from the long-term continuous monitoring test set, and in this case comprises data from 7 accelerometers corresponding to one side of the Z24 bridge. The first segment (8192 data points), with $f_s = 100$Hz, was used in the analysis. The priors of the Bayesian SSI model were left unchanged from those used in the previous numerical simulation (see Section \ref{sec: numsim - prior}) except for the precision hyperparameter $K_0$, which was changed to the identity providing some additional flexibility to the model.

\subsection{Single model order}

\begin{figure}[h!]
	\begin{subfigure}{\textwidth}
		\centering 
		\includegraphics[width=\textwidth]{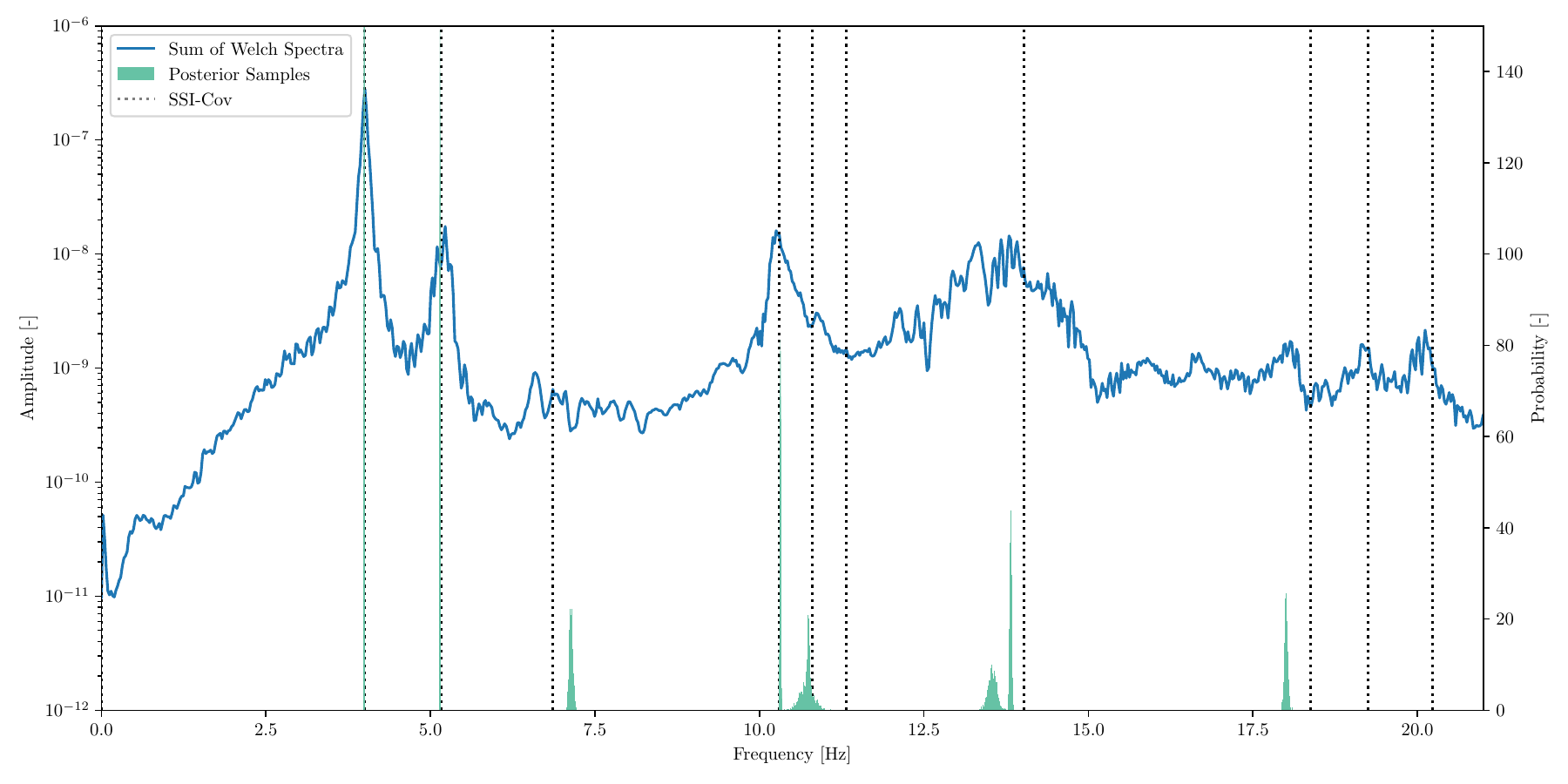}
		\caption{}
		\label{fig: z24 model order 10 big}
	\end{subfigure}\\
	\vspace{0.3cm}
	\begin{subfigure}{0.5\textwidth}
		\centering
		\includegraphics[width=\textwidth]{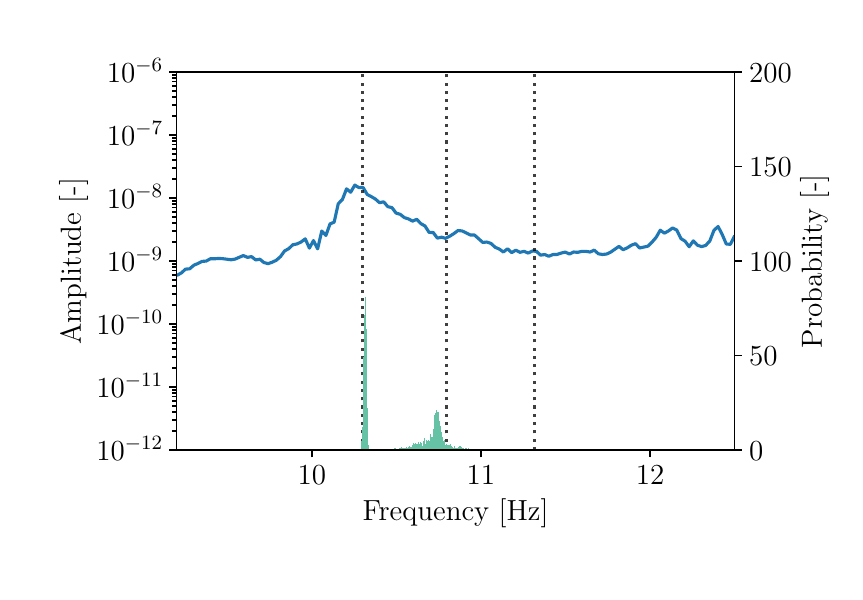}
		\caption{}
		\label{fig: z24 model order 10 zoomed 1}
	\end{subfigure}
	\hfill
	\begin{subfigure}{0.5\textwidth}
		\centering
		\includegraphics[width=\textwidth]{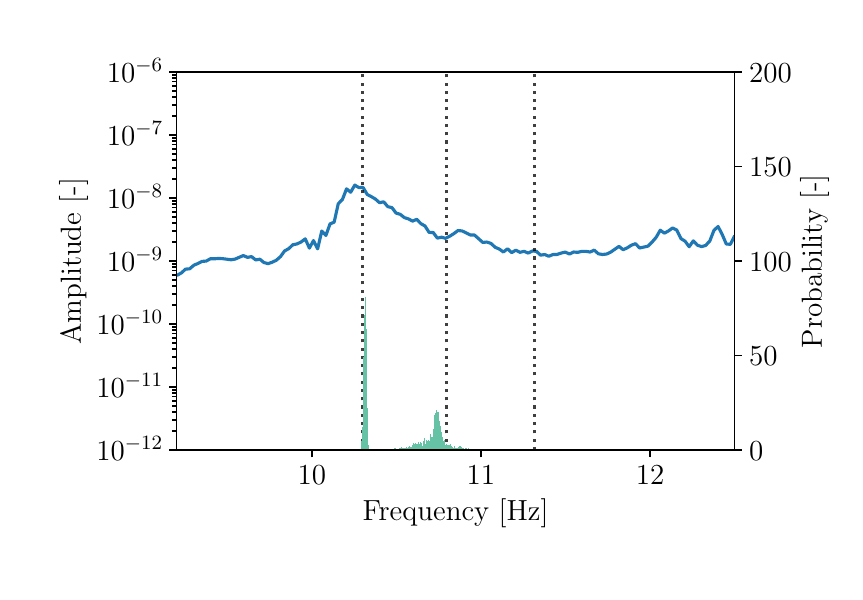}
		\caption{}
		\label{fig: z24 model order 10 zoomed 2}
	\end{subfigure}
	\caption{The sum of the Welch spectra of the Z24 data overlaid with the posteriors of the natural frequencies, represented by histograms of the natural frequency samples recovered using Bayesian SSI at a model order of 10 is given in (a). Regions of interest are shown in (b) and (c), with axes limits chosen to highlight the differences in the mean and variance of the histograms in different regions.}
	\label{fig: z24 model order 10}
\end{figure}

\begin{figure}[h!]
	\begin{subfigure}{\textwidth}
		\centering 
		\includegraphics[width=\textwidth]{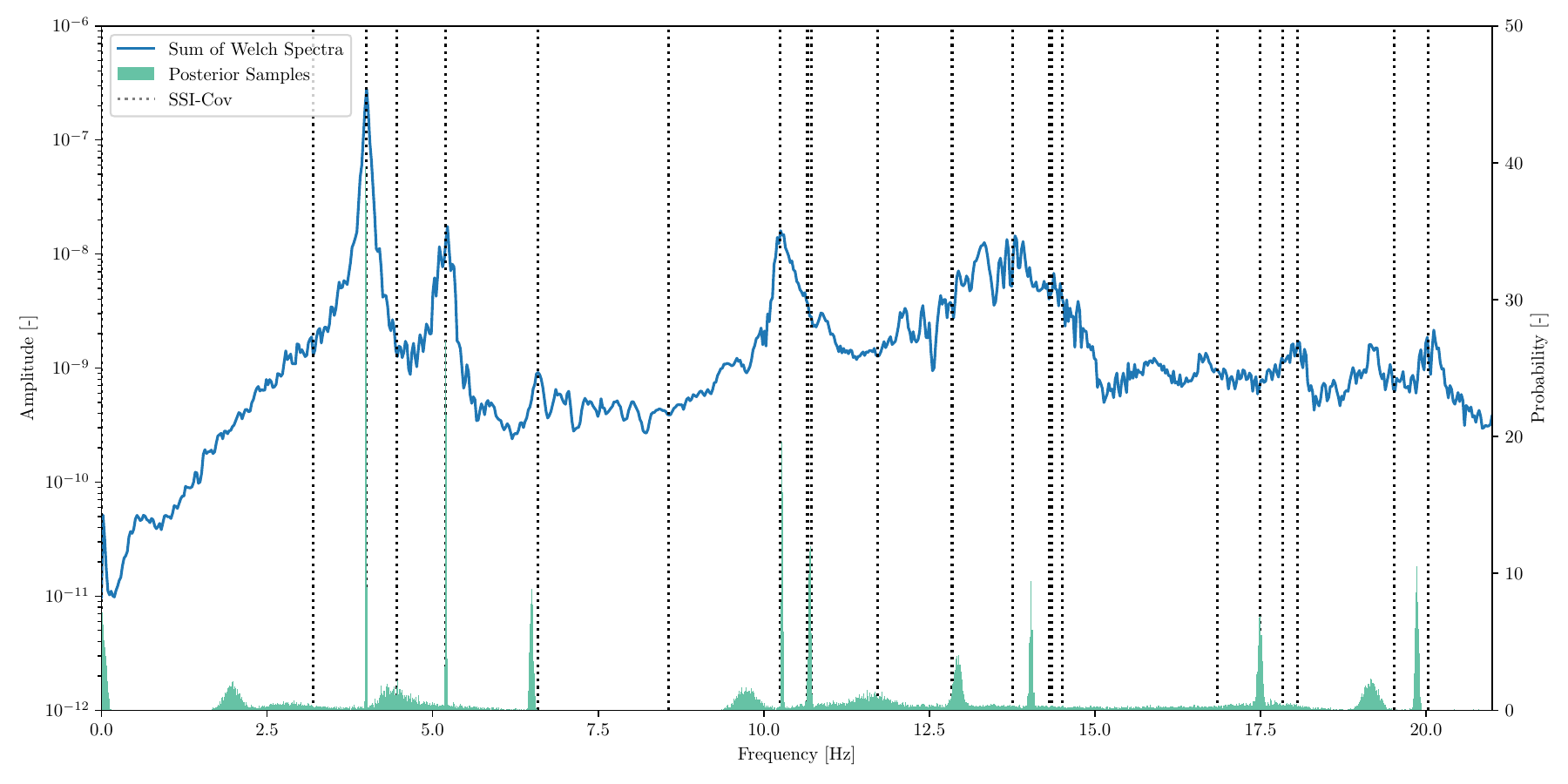}
		\caption{}
		\label{fig: z24 model order 30 big}	
	\end{subfigure}\\
	\vspace{0.3cm}
	\begin{subfigure}{0.5\textwidth}
		\centering
		\includegraphics[width=\textwidth]{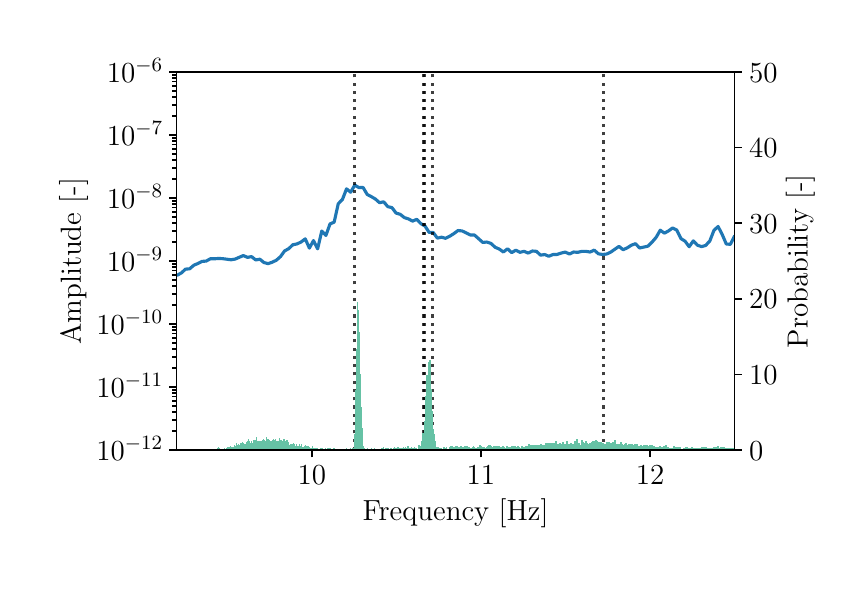}
		\caption{}
		\label{fig: z24 model order 30 zoomed 1}
	\end{subfigure}
	\hfill
	\begin{subfigure}{0.5\textwidth}
		\centering
		\includegraphics[width=\textwidth]{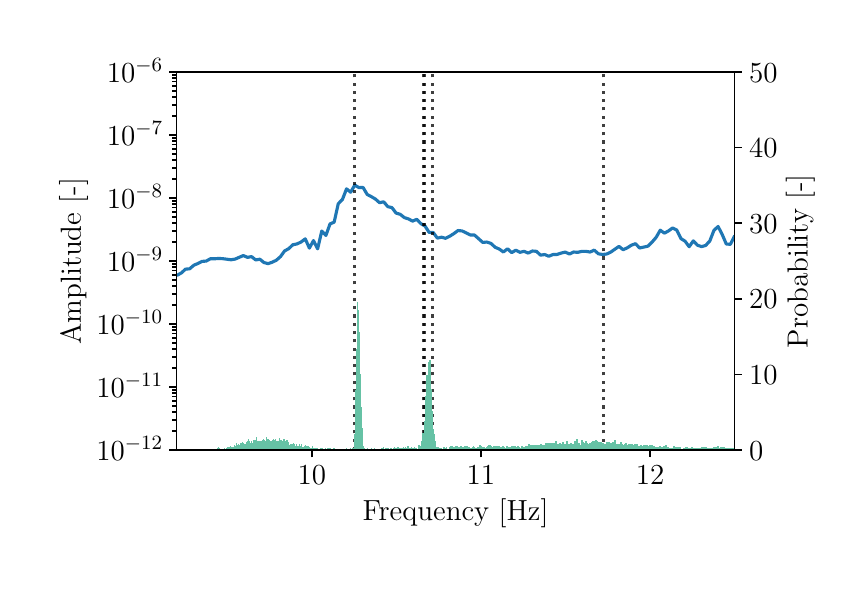}
		\caption{}
		\label{fig: z24 model order 30 zoomed 2}
	\end{subfigure}
	\caption{The average Welch spectrum of the Z24 data overlaid with the posteriors of the natural frequencies, represented by histograms of the natural frequency samples recovered by Bayesian SSI at a model order of 30, is given in (a). Regions of interest are shown in (b) and (c), with axes limits chosen to highlight the differences in the mean and variance of the histograms in different regions.}
	\label{fig: z24 model order 30}
\end{figure}

Conducting analysis at model orders of 10 and 30, chosen solely for demonstrative purposes, the posterior distributions of the modal characteristics were obtained using the variational implementation of Bayesian SSI. The posteriors over the natural frequencies, represented by histograms of the samples and overlaid on the sum of the Welch spectra in each channel, are shown in Figures \ref{fig: z24 model order 10} and \ref{fig: z24 model order 30}. 

It is evident from both Figure \ref{fig: z24 model order 10 big} and Figure \ref{fig: z24 model order 30 big} that histograms with means centred around the apparent natural frequencies (peaks in the Welch spectrum) demonstrate significantly lower variance than histograms corresponding to frequencies likely more spurious in nature. This lower variance can be seen more clearly in the enhanced Figures \ref{fig: z24 model order 10 zoomed 1}, \ref{fig: z24 model order 10 zoomed 2}, \ref{fig: z24 model order 30 zoomed 1} and \ref{fig: z24 model order 30 zoomed 2} and is most noticeable around 4 Hz, 5 Hz, 10-11 Hz. In contrast, larger variances are observed in regions where there is typically less evidence of a natural frequency. Focusing specifically on the case where the model order is 30, larger variances can be seen at 4.5 Hz, 9.75Hz and 11.75 Hz. This obvious variability in the variance for each identified frequency suggests that there is a higher probability or belief in some frequencies to describe the data over others.

\newpage
\subsection{Stabilisation diagram}

As the true number of modes is unknown, it is standard practice in OMA to construct a stabilisation diagram. Stabilisation diagrams are a tool used by the practitioner to determine the ``physical'' poles of a system, helping to decide the nature of the identified complex eigenvalues i.e. spurious or real in a physical context. The decision of which is often governed by some heuristic criteria on the stability or consistency of the modal parameters across a range of model orders.

\begin{figure}[h]
	\centering 
	\includegraphics[width=\textwidth]{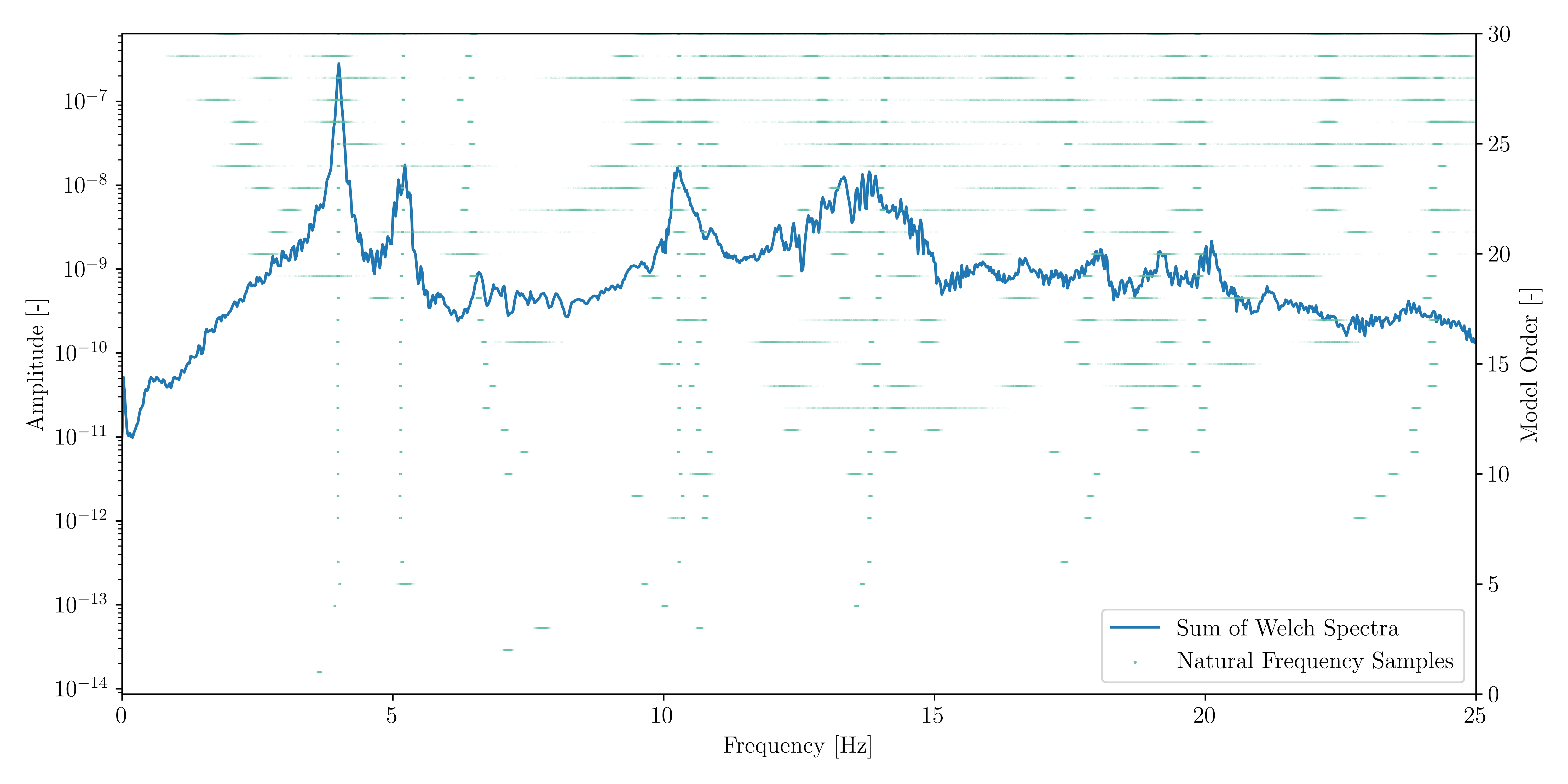}
	\caption{Stabilisation diagram of the Z24 bridge, constructed using the natural frequency samples obtained using Bayesian SSI at a range of model orders.}
	\label{fig: z24 stabilisation}
\end{figure}

After conducting Bayesian SSI at multiple model orders, the resulting stabilisation diagram for the Z24 is shown in Figure \ref{fig: z24 stabilisation}. The original 2-dimensional histogram representation of the posterior at single model orders (see Figures \ref{fig: z24 model order 10} and \ref{fig: z24 model order 30}) is ill-suited for the multi-order stabilisation diagram. Instead, samples are plotted with transparency such that darker regions indicate areas with a higher density of samples. The inclusion of uncertainty in stabilisation diagrams is not an entirely new concept, having already been represented, and proven beneficial in assisting model order selection \cite{priou2022}, although no agreed form has yet been defined. In this plot, non-conjugate poles have been removed, however no stability metrics have been defined, purely to observe the raw form of the stabilisation diagram.

Considering the distributions of the posteriors in Figure \ref{fig: z24 stabilisation}, one can clearly see that peaks in the Welch spectrum, typically indicative of a natural frequency, coincide with the posterior means and are consistent across multiple model orders. This is true for multiple peaks, most distinctly around 4 Hz, 5 Hz and 10-11 Hz. The variance of the distributions in these areas is also low across a broad range of model orders, especially when compared to more spurious frequency estimates away from natural frequencies. Thus indicating a higher level of confidence in the ability of certain frequencies to describe the dynamics. Another conspicuous observation is the general increase in variance at higher model orders, particularly in the more spurious estimates, as moving to higher model orders grant greater flexibility with numerous ways for the model to explain the data. 

\section{Conclusion}\label{sec:Conclusion}

In this work, a new Bayesian perspective on OMA has been presented. Using the probabilistic interpretation of SSI-Cov, it has been shown that prior knowledge can be incorporated over the model parameters to form a Bayesian SSI algorithm capable of recovering posterior distributions over modal characteristics of interest. This new approach was presented with two possible methods of inference: Markov chain Monte Carlo and variational Bayes. These two forms of Bayesian SSI were then benchmarked using a simulated case study. The prior distributions over the modal parameters were shown, followed by the recovered posteriors; which demonstrated good agreement with one another and convergence of the posterior mean towards the conventional SSI result. The effects of data length on the recovered posteriors was also briefly explored. It was shown that, with increasing data length, the posterior mean trends toward the truth whilst the variance decreases as expected. Finally, the practicality and applicability of the algorithm was then illustrated using data obtained from an in-service structure, the Z24 bridge. Recovery of the modal parameters and associated posterior uncertainty was shown at single model orders and in a stabilisation diagram using the natural frequencies estimates. The results showed that posteriors with means centred around the apparent natural frequencies display much lower variance when compared to more spurious frequency estimates, further away from the natural frequencies. 

\subsection{Future Work}

As is typical with Bayesian UQ, the choice and appropriate specification of the priors is an important task. Priors should reflect ones initial belief, yet they can also be chosen tactically to ensure faster convergence to the posterior under limited data or for a desired conjugacy with the likelihood. In the case of Bayesian SSI, prior distributions are not placed over the modal properties but instead over the columns of the weight matrices (the observability and controllability). This choice of prior setting remains unchanged from the initial definition of Bayesian CCA \cite{wang2007,klami2007} but may not be the most suitable for this chosen application. At present, the chosen prior can result in posterior densities that include negative estimates for the damping ratio. Negative values of damping would be consistent with non-physical attributes to a supposedly stable dynamic system. Nevertheless, negative estimates for the damping ratio can numerically occur in conventional SSI, meaning posterior densities containing negative damping are plausible given the current model choice. It is interesting to imagine if priors over the model parameters could be found, such that the modal properties are constrained to be physically meaningful. This possible prior specification is to be explored in future work. 

In other aspects, it is hoped that the methodologies introduced in this work will enable researchers to tackle existing OMA research challenges and explore the many capabilities of this now Bayesian SSI algorithm; including alternative noise models, the effect of alternative prior specification, decision-making tasks, computationally efficient implementations, and inclusion in existing frameworks.

In summary, this work has shown that posterior uncertainty over the modal parameters has significant benefits. The availability of this uncertainty information means it can be propagated to assist in the solution of other important research challenges in OMA, such as model order selection, automatic OMA and sensor placement. In SHM frameworks, one could foresee the inclusion of modal posterior uncertainties as a valuable addition to damage assessment and decision-making; providing new information one could envisage altering decision boundaries and classification models. Going forward, the authors would promote the Bayesian SSI technique presented in this work as a powerful building block for use in modal analysis and beyond.

\section*{Acknowledgements}
The authors gratefully acknowledge the support of the Engineering and Physical Sciences Research Council (EPSRC), UK through grant numbers EP/W002140/1 and EP/W005816/1. For the purpose of open access, the authors have applied a Creative Commons Attribution (CC-BY) license to any Author Accepted Manuscript version arising.

\section*{CRediT authorship contribution statement}

\textbf{Brandon J. O'Connell}: Conceptualization, Writing - original draft, Visualization, Validation, Methodology, Software. \textbf{Max D. Champneys}: Writing - review \& editing, Methodology. \textbf{Timothy J. Rogers}: Writing - review \& editing, Supervision, Conceptualization, Methodology, Software, Funding acquisition.

\appendix

\section{Canonical-variate weighted, covariance-driven stochastic subspace identification} \label{A: SSI-Cov deriv}

For the reader's benefit, a concise description of canonical-variate weighted SSI-Cov algorithm for an output-only case is given here. This derivation follows descriptions for the balanced (canonical-variate weighted) algorithm given in Katayama \cite{katayama2005} and Van Overschee and De Moor \cite{vanoverschee1996}. The reader is directed towards the aforementioned texts for a complete exposition.
	
Consider an $r^{\mathrm{th}}$ order discrete state space model of a linear dynamic system, equivalent to a mechanical system with $n_{\mathrm{dof}}$ degrees of freedom, such that $r = 2 n_{\mathrm{dof}}$, in the form

\begin{equation}
	\begin{aligned}
	\vec{x}_{k+1} &= \vec{A}_d \vec{x}_k + w_k \\
	\vec{y}_k &= \vec{C} \vec{x}_k + v_k
	\end{aligned}
	\quad , \quad
	\mathbb{E} \left[ \begin{bmatrix} w_q \\ v_q \end{bmatrix} \begin{bmatrix} w^{\trans}_s & v^{\trans}_s \end{bmatrix} \right] = \begin{bmatrix} \vec{Q} & \vec{S} \\ \vec{S}^{\trans} & \vec{R} \end{bmatrix} \delta_{qs}
	\label{eq: State Space Model}
\end{equation}

where $\vec{y}_k \in \mathbb{R}^{l}$ is the output vector at discrete time step $k$, $\vec{x}_k \in \mathbb{R}^{p}$ is the internal state vector, $\vec{A}_d \in \mathbb{R}^{p \times p}$ is the discrete state matrix such that $\vec{A}_d = \mathrm{expm}(\vec{A}_c \Delta t)$ where $\vec{A}_c$ is the continuous state matrix, $\Delta t$ is the sampling time and $\vec{C}$ is the output matrix. The terms $w_k \in \mathbb{R}^{p}$ and $v_k \in \mathbb{R}^{l}$ are samples of the process noise and measurement noise respectively, $\mathbb{E}[\cdot]$ denotes the expectation and $\delta_{qs}$ is the Kronecker delta for any two samples in time $q$ and $s$. The process and measurement noise are both assumed to be stationary, white noise Gaussian with zero mean and some covariance given by the second part of Eq.(\ref{eq: State Space Model}).

	
Given output measurements from this stationary process, based on $l$ measurement channels, the data can be arranged into a block Hankel matrix $\mathcal{H} \in \mathbb{R}^{2lj \times N}$ to form     

\begin{equation}
   \mathcal{H} = \vec{Y}_{0|2j-1}  \ \ =  \ \ \begin{bmatrix} \vec{Y}_{0|j-1} \\ \vec{Y}_{j|2j-1}\end{bmatrix} \ \ = \ \ \begin{bmatrix} \vec{Y}_p \\ \vec{Y}_f \end{bmatrix} 
\end{equation}

\noindent with $2j$ block rows and $N$ columns, with a block consisting of $l$ rows and where $j>0$ and $N$ is sufficiently large (i.e much larger than $2lj$) and where $j > r$ and the number of columns of block matrices is $N$. The resultant cross-covariance matrix of the future $\vec{Y}_f$ with the past $\vec{Y}_p$ is therefore given by

\begin{equation}
	\tilde{\mat{\Sigma}} \ \ = \ \ \dfrac{1}{N}\begin{bmatrix} \vec{Y}_p \\ \vec{Y}_f \end{bmatrix} \begin{bmatrix} \vec{Y}_p^{\trans} & \vec{Y}_f^{\trans} \end{bmatrix} \ \ = \ \ \begin{bmatrix} \vec{\Sigma}_{pp} & \vec{\Sigma}_{pf} \\ \vec{\Sigma}_{fp} & \vec{\Sigma}_{ff} \end{bmatrix} 
\end{equation}
    
\noindent where $\vec{\Sigma}_{pf}$ and $\vec{\Sigma}_{fp}$ are block cross-covariance matrices, and $\vec{\Sigma}_{ff}$, $\vec{\Sigma}_{pp}$ are block auto-covariance matrices respectively. The canonical correlations $\vec{\Lambda} = \mathrm{diag}(\lambda_1, \cdots, \lambda_r)$ between the future and past are the singular values \cite{katayama2005}, obtained through the SVD of the following matrix

	\begin{equation}
		\vec{\Sigma}_{ff}^{-1/2}\vec{\Sigma}_{fp}\vec{\Sigma}_{pp}^{-\trans/2} \ \ = \ \ \vec{V}_1 \vec{\Lambda} \vec{V}_2^{\trans} \ \ \backsimeq \ \ \breve{\vec{V}}_1 \breve{\vec{\Lambda}} \breve{\vec{V}}_2^{\trans}
		\label{eq:Normalised Covariance 1}
	\end{equation}

\noindent where $\vec{\Sigma}_{ff}^{1/2}\vec{\Sigma}_{ff}^{\trans/2} = \vec{\Sigma}_{ff}$, such that, 

	\begin{equation}
		\vec{\Sigma}_{fp} \ \ \backsimeq \ \  \vec{\Sigma}_{ff}^{1/2} \breve{\vec{V}}_1 \breve{\vec{\Lambda}} \breve{\vec{V}}_2^{\trans} \vec{\Sigma}_{pp}^{\trans/2}
		\label{eq:Normalised Covariance 2}
	\end{equation}

\noindent where $\vec{V_1}$ and $\vec{V}_2$ are the left and right singular vectors, respectively and $\breve{\vec{\Lambda}}$ neglects sufficiently small singular values (or canonical correlations) in $\vec{\Lambda}$ such that the resultant state vector has the dimension $d = \mathrm{dim}(\breve{\vec{\Lambda}})$. The cross-covariance matrix, $\vec{\Sigma}_{pf}$, can be decomposed into the corresponding extended observability $(\mathcal{O})$ and controllability $(\mathcal{C})$ matrices using $\vec{\Sigma}_{fp} = \mathcal{O}\mathcal{C}$ such that
	\begin{equation}
		\mathcal{O}  \ \ = \ \ \vec{\Sigma}_{ff}^{1/2}\breve{\vec{V}}_1\breve{\vec{\Lambda}}^{1/2} \ \ \ , \ \ \ \ \mathcal{C} \ \ = \ \ \breve{\vec{\Lambda}}^{1/2}\breve{\vec{V}}_2^{\trans}\vec{\Sigma}_{pp}^{\trans/2}
		\label{eq:ObservReach}
	\end{equation}

 with rank($\mathcal{O}$) = rank($\mathcal{C}$) = $d$. Note in the context of output only SSI, the controllability has little meaning. The extended observability and controllability matrices can then be used to recover the state $\vec{A}$ and output $\vec{C}$ matrices, and consequently the modal properties, in the usual manner for operational modal analysis (see Section 3.3 in \cite{reynders2008} and also \cite{ewins2000}). The method described here for recovering the system matrices is one of many. See stochastic balanced realisation algorithm B, Chapter 8 in Katayama \cite{katayama2005}.

\newpage
\section{Bayesian CCA - Gibbs Sampler Derivation}\label{A: gibbs}

This appendix provides a full derivation for the update equations provided in Algorithm \ref{al: BayCCA Gibbs algorithm}, aligning with the Bayesian CCA formulation presented in Klami and Kaski \cite{klami2007} but without the sparsity prior and with the inclusion of generic priors. 

Given the model defined in Figure \ref{fig: bayesian cca graphical model}, the joint likelihood of Bayesian CCA can be defined as

\begin{equation}
	\mathcal{L} = p(\vec{x}, \vec{z}, \vec{W}, \vec{\Sigma}, \vec{\mu}) = p(\vec{x} | \vec{z}, \vec{W}, \vec{\Sigma}, \vec{\mu}) p(\vec{z}) p(\vec{\Sigma}) p(\vec{\mu}) p(\vec{W})
\end{equation}

Assuming independent columns of W, the joint can also be defined as

\begin{equation}
	\mathcal{L} = p(\vec{x} | \vec{z}, \vec{W}, \vec{\Sigma}, \vec{\mu}) p(\vec{z}) p(\vec{\Sigma}) p(\vec{\mu}) \prod_{i=1}^{d} p(\vec{w}_i) 
\end{equation} 

Taking the natural log of the joint likelihood, the individual components of the log joint are given by 

\begin{equation}
	\ln \mathcal{L}_{p(\vec{x} | \vec{z}, \vec{W}, \vec{\Sigma}, \vec{\mu})} \propto -\dfrac{N}{2}\ln|{\mat{\Sigma}}| - \hlf \sum_{n=1}^{N} (\vec{x}_n - (\vec{W}\vec{z}_n + \vec{\mu}))^\trans \vec{\Sigma}^{-1} (\vec{x}_n -(\vec{W}\vec{z}_n + \vec{\mu})) + \mathrm{const}
\end{equation}

\begin{equation}
	\ln \mathcal{L}_{p(\vec{z})} \propto -\dfrac{N}{2}\ln|{\ident}| - \hlf \sum_{n=1}^{N} (\vec{z}_n - 0)^\trans \ident^{-1} (\vec{z}_n - 0) = - \hlf \sum_{n=1}^{N} \vec{z}_n^{\trans}\vec{z}_n + \mathrm{const}\\
\end{equation}

\begin{equation}
	\ln \mathcal{L}_{\vec{\Sigma}} \propto \dfrac{\nu_0}{2}\ln|\vec{K}_0| - \dfrac{\nu_0 D}{2} \ln(2) - \ln \left (\Gamma \left (\dfrac{\nu_0}{2}\right ) \right ) - \dfrac{\nu_0 + D + 1}{2}\ln|\vec{\Sigma}| - \hlf \trace{\vec{K}_0\vec{\Sigma}^{-1}} + \mathrm{const}
\end{equation}

\begin{equation}
	\ln \mathcal{L}_{p(\vec{\mu})} \propto -\dfrac{1}{2}\ln|\vec{\Sigma}_{\mu}| - \hlf \sum_{n=1}^{N} (\vec{\mu} - \vec{\mu}_{\mu})^\trans \vec{\Sigma}_{\mu}^{-1} (\vec{\mu} - \vec{\mu}_{\mu}) + \mathrm{const}\\
\end{equation}

\begin{equation}
	\ln \mathcal{L}_{p(\vec{w}_i)} \propto -\hlf \ln|\vec{\Sigma}_{\vec{w}_i}| - \hlf \sum_{n=1}^{N} (\vec{w}_i - \vec{\mu}_{\vec{w}_i})^\trans \vec{\Sigma}_{\vec{w}_i}^{-1} (\vec{w}_i - \vec{\mu}_{\vec{w}_i}) + \mathrm{const}\\
\end{equation}
where 
\begin{equation}
	\prod_{i=1}^{d} p(\vec{w}_i) = \exp\left\{\sum^{d}_{i=1} \ln p(\vec{w}_i)\right\}
\end{equation}

Using the standard theory of Gibbs MCMC sampling the update equations for the model parameters can be constructed. The premise of Gibbs sampling is to sample each variable in turn, conditioned on the values of the other variables in the joint distribution to obtain a sample from the conditional distribution that is sought. This  can be achieved by considering the log of the joint likelihood with respect to the parameter of interest, establishing the new conditional distribution, and evaluating that function given the current estimates for all the necessary model parameters.

The following subsections summarise the derivation of the individual updates. Even though a specified order is given in Algorithm \ref{al: BayCCA Gibbs algorithm}, the nature of Gibbs sampling means that sampling from the conditionals can be conducted in any order. As such, the equations here are written without specifying which parameters are current or previous estimates, to simplify notation for clarity. 

\subsection{Sample $\vec{\Sigma}$}

Collecting the terms of the log joint pertaining to $\vec{\Sigma}$

\begin{multline}
	\ln{p(\vec{\Sigma}^{(\tau + 1)})} \propto -\dfrac{N}{2}\ln|\vec{\Sigma}| - \hlf\sum_{n=1}^N (\vec{x}_n - (\vec{W}\vec{z}_n + \vec{\mu}))^\trans \vec{\Sigma}^{-1} (\vec{x}_n - (\vec{W}\vec{z}_n + \vec{\mu}))  \\ - \dfrac{\nu_0 + D + 1}{2}\ln|\vec{\Sigma}| - \hlf \trace{\vec{K}_0\vec{\Sigma}^{-1}}
\end{multline}

Using $\vec{x}^{\trans}\vec{A}\vec{x} = \mathrm{trace}(\vec{A}\vec{x}\vec{x}^{\trans})$,

\begin{multline}
	\ln{p(\vec{\Sigma}^{(\tau + 1)})} \propto -\dfrac{N}{2}\ln|\vec{\Sigma}| - \hlf \trace{\vec{\Sigma}^{-1} \sum_{n=1}^N (\vec{x}_n - \vec{\mu} - \vec{W}\vec{z}_n)(\vec{x}_n - \vec{\mu} - \vec{W}\vec{z}_n)^\trans}  \\ - \dfrac{\nu_0 + D + 1}{2}\ln|\vec{\Sigma}| - \hlf \trace{\vec{K}_0\vec{\Sigma}^{-1}}
\end{multline}

Combining terms, the familiar inverse Wishart form can be obtained

\begin{equation}
	\ln{p(\vec{\Sigma}^{(\tau + 1)})} \propto - \dfrac{\nu_0 + N + D + 1}{2}\ln|\vec{\Sigma}| - \hlf \trace{\left (\vec{K}_0 + \sum_{n=1}^N (\vec{x}_n - \vec{\mu} - \vec{W}\vec{z}_n ) (\vec{x}_n - \vec{\mu} - \vec{W}\vec{z}_n)^\trans \right ) \vec{\Sigma}^{-1} }
\end{equation}

Therefore, a new sample of $\vec{\Sigma}$ can be drawn from the found conditional distribution

\begin{equation}
	\vec{\Sigma}^{(\tau + 1)} \sim \iwishdistr{\vec{K}_0 + \vec{K}}{\nu_0 + N} 
\end{equation}
where 
\begin{equation}
	\vec{K} = \sum_{n=1}^N (\vec{x}_n - \vec{\mu} - \vec{W}\vec{z}_n ) (\vec{x}_n - \vec{\mu} - \vec{W}\vec{z}_n)^\trans
\end{equation}

\subsection{Sample $\vec{\mu}$}

Collecting the terms of the log joint pertaining to $\vec{\mu}$

\begin{multline}
	\ln{p(\vec{\mu}^{(\tau + 1)})} \propto - \hlf\sum_{n=1}^N \left \{(\vec{x}_n - (\vec{W}\vec{z}_n + \vec{\mu}))^\trans \vec{\Sigma}^{-1} (\vec{x}_n - (\vec{W}\vec{z}_n + \vec{\mu})) \right \} - \hlf (\vec{\mu} - \vec{\mu}_{\mu})^\trans \vec{\Sigma}_{\mu}^{-1} (\vec{\mu} - \vec{\mu}_{\mu})
\end{multline}



Expanding and ignoring terms not containing $\vec{\mu}$,

\begin{multline}
	\ln{p(\vec{\mu}^{(\tau + 1)})} \propto - \hlf\sum_{n=1}^N \left \{ \vec{\mu}^\trans \vec{\Sigma}^{-1} \vec{\mu} - \vec{\mu}^\trans \vec{\Sigma}^{-1}(\vec{x}_n - \vec{W}\vec{z}_n) - (\vec{x}_n - \vec{W}\vec{z}_n)^\trans \vec{\Sigma}^{-1}\vec{\mu} \right \}   \\ - \hlf (\vec{\mu}^\trans \vec{\Sigma}_{\mu}^{-1}\vec{\mu} - \vec{\mu}_{\mu}^\trans \vec{\Sigma}_{\mu}^{-1} \vec{\mu} - \vec{\mu}^\trans \vec{\Sigma}_{\mu}^{-1} \vec{\mu}_{\mu} )
\end{multline}



setting $\hat{\vec{\Sigma}}^{-1}_{\mu} = N\vec{\Sigma}^{-1} + \vec{\Sigma}_{\mu}^{-1}$, and in the knowledge that $\ident = \hat{\vec{\Sigma}}^{-1}_{\mu}\hat{\vec{\Sigma}}_{\mu}$, the log form of a Gaussian can be reached

\begin{multline}
	\ln{p(\vec{\mu}^{(\tau + 1)})} \propto - \hlf  \left \{ \vec{\mu}^\trans \hat{\vec{\Sigma}}^{-1}_{\mu} \vec{\mu} - \vec{\mu}^\trans \hat{\vec{\Sigma}}^{-1}_{\mu} \hat{\vec{\Sigma}}_{\mu} (\vec{\Sigma}^{-1} \sum^N_{n=1} (\vec{x}_n - \vec{W}\vec{z}_n) + \vec{\Sigma}_{\mu}^{-1} \vec{\mu}_{\mu}) \right . \\ \left . -  \left ( \sum^N_{n=1} (\vec{x}_n - \vec{W}\vec{z}_n)^{\trans}\vec{\Sigma}^{-1}  +  \vec{\mu}_{\mu}\vec{\Sigma}_{\mu}^{-1} \right ) \vec{\mu}^\trans \right \}
\end{multline}

Therefore, a new sample of $\vec{\mu}$ can be drawn from the conditional distribution defined by
\begin{equation}
	\vec{\mu}^{(\tau + 1)} \sim \gaussdistr{\hat{\vec{\mu}}_{\mu}}{\hat{\vec{\Sigma}}_{\mu}} 
\end{equation}
where 
\begin{equation}
	\hat{\vec{\Sigma}}_{\mu} = (N\vec{\Sigma}^{-1} + \vec{\Sigma}_{\mu}^{-1})^{-1}
\end{equation}
and 
\begin{equation}
	\hat{\vec{\mu}}_{\mu} = \hat{\vec{\Sigma}}_{\mu}(\vec{\Sigma}^{-1} \sum_{n=1}^N (\vec{x}_n - \vec{W}\vec{z}_n) + \vec{\Sigma}_{\mu}^{-1}\vec{\mu}_{\mu})
\end{equation}

\subsection{Sample $\vec{w}_i$}

Collecting the terms of the log joint pertaining to $\vec{W}$

\begin{multline}
	\ln{p(\vec{w}_i^{(\tau + 1)})} \propto - \hlf\sum_{n=1}^N \left \{(\vec{x}_n - (\vec{W}\vec{z}_n + \vec{\mu}))^\trans \vec{\Sigma}^{-1} (\vec{x}_n - (\vec{W}\vec{z}_n + \vec{\mu})) \right \}  \\ - \hlf \sum^{d}_{i=1} \left \{ (\vec{w}_i - \vec{\mu}_{\vec{w}_i})^\trans \vec{\Sigma}_{\vec{w}_i}^{-1} (\vec{w}_i - \vec{\mu}_{\vec{w}_i}) \right \}
\end{multline}

As the intention is to update only one column of $\vec{W}$ at a time, i.e the independent column $\vec{w}_i$, the above equation can be separated as follows

\begin{multline}
	\ln{p(\vec{w}_i^{(\tau + 1)})} \propto - \hlf\sum_{n=1}^N \left \{(\vec{x}_n - \vec{\mu} - \vec{w}_{\_i}\vec{z}_{\_i,n} - \vec{w}_{i}\vec{z}_{i,n})^\trans \vec{\Sigma}^{-1} (\vec{x}_n - \vec{\mu} - \vec{w}_{\_i}\vec{z}_{\_i,n} - \vec{w}_{i}\vec{z}_{i,n}) \right \}  \\ - \hlf (\vec{w}_i - \vec{\mu}_{\vec{w}_i})^\trans \vec{\Sigma}_{\vec{w}_i}^{-1} (\vec{w}_i - \vec{\mu}_{\vec{w}_i})
\end{multline}

where the negative indices notation, $\cdot_{\_i}$, refers to every column/row of except the one specified. Letting $\tilde{\vec{x}}_n  = \vec{x}_n - \vec{\mu} - \vec{w}_{\_i}\vec{z}_{\_i,n}$, this equation can be simplified to

\begin{equation}
	\ln{p(\vec{w}_i^{(\tau + 1)})} \propto - \hlf\sum_{n=1}^N \left \{(\tilde{\vec{x}}_n - \vec{w}_{i}\vec{z}_{i,n})^\trans \vec{\Sigma}^{-1} (\tilde{\vec{x}}_n - \vec{w}_{i}\vec{z}_{i,n}) \right \} - \hlf (\vec{w}_i - \vec{\mu}_{\vec{w}_i})^\trans \vec{\Sigma}_{\vec{w}_i}^{-1} (\vec{w}_i - \vec{\mu}_{\vec{w}_i})
\end{equation}

and expanded to give

\begin{multline}
	\ln{p(\vec{w}_i^{(\tau + 1)})} \propto - \hlf\sum_{n=1}^N \left \{\tilde{\vec{x}}_n^{\trans} \vec{\Sigma}^{-1} \tilde{\vec{x}}_n - (\vec{w}_{i}\vec{z}_{i,n})^\trans \vec{\Sigma}^{-1} \tilde{\vec{x}}_n - \tilde{\vec{x}}_n^\trans \vec{\Sigma}^{-1} \vec{w}_{i}\vec{z}_{i,n} + (\vec{w}_{i}\vec{z}_{i,n})^\trans \vec{\Sigma}^{-1} \vec{w}_{i}\vec{z}_{i,n} \right \}  \\ - \hlf (\vec{w}_i^\trans \vec{\Sigma}_{\vec{w}_i}^{-1} \vec{w}_i - \vec{w}_i^\trans \vec{\Sigma}_{\vec{w}_i}^{-1} \vec{\mu}_{\vec{w}_i} -  \vec{\mu}_{\vec{w}_i}^\trans \vec{\Sigma}_{\vec{w}_i}^{-1} \vec{w}_i + \vec{\mu}_{\vec{w}_i}^\trans \vec{\Sigma}_{\vec{w}_i}^{-1} \vec{\mu}_{\vec{w}_i})
\end{multline}



As $\vec{z}_{i,n}$ is a scalar, it can be manipulated as such, rearranging to give

\begin{multline}
	\ln{p(\vec{w}_i^{(\tau + 1)})} \propto - \hlf \left \{ \sum_{n=1}^N \{ \vec{z}_{i,n}\vec{z}_{i,n} \} \vec{w}_{i}^\trans \vec{\Sigma}^{-1} \vec{w}_{i} - \vec{w}_i^\trans \vec{\Sigma}_{\vec{w}_i}^{-1} \vec{w}_i - \sum_{n=1}^N \vec{z}_{i,n}\vec{w}_{i}^\trans \vec{\Sigma}^{-1} \tilde{\vec{x}}_n - \sum_{n=1}^N \vec{z}_{i,n}\tilde{\vec{x}}_n^\trans \vec{\Sigma}^{-1} \vec{w}_{i} \right .  \\ \left .-  \vec{w}_i^\trans \vec{\Sigma}_{\vec{w}_i}^{-1} \vec{\mu}_{\vec{w}_i} -  \vec{\mu}_{\vec{w}_i}^\trans \vec{\Sigma}_{\vec{w}_i}^{-1} \vec{w}_i \right \}
\end{multline}


setting $\hat{\vec{\Sigma}}^{-1}_{\vec{w}_i} = \sum_{n=1}^N \{ \vec{z}_{i,n}\vec{z}_{i,n} \} \vec{\Sigma}^{-1} - \vec{\Sigma}_{\vec{w}_i}^{-1}$, and in the knowledge that $\ident = \hat{\vec{\Sigma}}^{-1}_{\vec{w}_i}\hat{\vec{\Sigma}}_{\vec{w}_i}$, the following log normal form can be reached

\begin{multline}
	\ln{p(\vec{w}_i^{(\tau + 1)})} \propto - \hlf \left \{\vec{w}_{i}^\trans \hat{\vec{\Sigma}}^{-1}_{\vec{w}_i} - \vec{w}_{i}^\trans \hat{\vec{\Sigma}}^{-1}_{\vec{w}_i} \hat{\vec{\Sigma}}_{\vec{w}_i} \left ( \sum_{n=1}^N \vec{z}_{i,n} \vec{\Sigma}^{-1} \tilde{\vec{x}}_n + \vec{\Sigma}_{\vec{w}_i}^{-1} \vec{\mu}_{\vec{w}_i} \right ) \right . \\ \left .- \left (\sum_{n=1}^N \vec{z}_{i,n}\tilde{\vec{x}}_n^\trans \vec{\Sigma}^{-1} + \vec{\mu}_{\vec{w}_i}^\trans \vec{\Sigma}_{\vec{w}_i}^{-1} \right ) \hat{\vec{\Sigma}}_{\vec{w}_i} \hat{\vec{\Sigma}}^{-1}_{\vec{w}_i} \vec{w}_{i} \right \}
\end{multline}

Therefore, a new sample of $\vec{w}_i$ can be drawn from the conditional distribution defined by
\begin{equation}
	\vec{w}_i^{(\tau + 1)} \propto \gaussdistr{\hat{\vec{\mu}}_{\vec{w}_i}}{\hat{\vec{\Sigma}}_{\vec{w}_i}} 
\end{equation}
where 
\begin{equation}
	\hat{\vec{\Sigma}}_{\vec{w}_i} = \left (\sum_{n=1}^N \{ \vec{z}_{i,n}\vec{z}_{i,n} \} \vec{\Sigma}^{-1} + \vec{\Sigma}_{\vec{w}_i}^{-1} \right )^{-1}
\end{equation}
and 
\begin{equation}
	\hat{\vec{\mu}}_{\vec{w}_i} = \hat{\vec{\Sigma}}_{\vec{w}_i}(\vec{\Sigma}^{-1} \sum_{n=1}^N \vec{z}_{i,n}\tilde{\vec{x}}_n + \vec{\Sigma}_{\vec{w}_i}^{-1}\vec{\mu}_{\vec{w}_i})
\end{equation}

Since each column of $\mat{W}$ is sampled independently, two methods of sampling can be used. Either: new columns are used in the following column updates, or the columns of the weight matrix are all sampled using the previous value for $\mat{W}$ before finally updating the full matrix.

\subsection{Sample $\vec{z}$}

Collecting the terms of the log joint pertaining to $\vec{z}$

\begin{equation}
	\ln{p(\vec{z}_n^{(\tau + 1)})} \propto - \hlf\sum_{n=1}^N \left \{(\vec{x}_n - (\vec{W}\vec{z}_n + \vec{\mu}))^\trans \vec{\Sigma}^{-1} (\vec{x}_n - (\vec{W}\vec{z}_n + \vec{\mu})) \right \} - \hlf \sum_{n=1}^N\vec{z}_n^\trans \vec{z}_n
\end{equation}

which can be written as

\begin{equation}
	\ln{p(\vec{z}_n^{(\tau + 1)})} \propto - \hlf\sum_{n=1}^N \left \{(\vec{x}_n - \vec{\mu} - \vec{W}\vec{z}_n)^\trans \vec{\Sigma}^{-1} (\vec{x}_n - \vec{\mu} - \vec{W}\vec{z}_n) + \vec{z}_n^\trans \vec{\ident}\vec{z}_n \right \}
\end{equation}



Expanding and ignoring any terms not containing $\vec{z}_n$,

\begin{multline}
	\ln{p(\vec{z}_n^{(\tau + 1)})} \propto - \hlf\sum_{n=1}^N \left \{- \vec{x}_n^\trans\vec{\Sigma}^{-1}\vec{W}\vec{z}_n + \vec{\mu}^\trans\vec{\Sigma}^{-1}\vec{W}\vec{z}_n - \vec{z}_n^\trans\vec{W}^\trans\vec{\Sigma}^{-1}\vec{x}_n + \vec{z}_n^\trans\vec{W}^\trans\vec{\Sigma}^{-1}\vec{\mu}  \right . \\ \left . + \vec{z}_n^\trans\vec{W}^\trans\vec{\Sigma}^{-1}\vec{W}\vec{z}_n + \vec{z}_n^\trans \vec{\ident} \vec{z}_n \right \}
\end{multline}



letting $\hat{\vec{\Sigma}}^{-1}_{\vec{z}} = (\vec{W}^\trans\vec{\Sigma}^{-1}\vec{W} + \vec{\ident})$, and in the knowledge that $\ident = \hat{\vec{\Sigma}}^{-1}_{\vec{z}}\hat{\vec{\Sigma}}_{\vec{z}}$, the log form of a Gaussian can be reached

\begin{equation}
	\ln{p(\vec{z}_n^{(\tau + 1)})} \propto - \hlf\sum_{n=1}^N \left \{\vec{z}_n^\trans \hat{\vec{\Sigma}}_{\vec{z}}\vec{z}_n - (\vec{x}_n - \vec{\mu})^\trans\vec{\Sigma}^{-1}\vec{W}\hat{\vec{\Sigma}}_{\vec{z}}\hat{\vec{\Sigma}}^{-1}_{\vec{z}}\vec{z}_n - \vec{z}_n^\trans\hat{\vec{\Sigma}}^{-1}_{\vec{z}}\hat{\vec{\Sigma}}_{\vec{z}}\vec{W}^\trans\vec{\Sigma}^{-1}(\vec{x}_n - \vec{\mu}) \right \}
\end{equation}

Therefore, a new sample of $\vec{z}$ can be drawn from the conditional distribution defined by

\begin{equation}
	\vec{z}_n^{(\tau + 1)} \sim \gaussdistr{\hat{\vec{\mu}}_{\vec{z}}}{\hat{\vec{\Sigma}}_{\vec{z}}} 
\end{equation}
where 
\begin{equation}
	\hat{\vec{\Sigma}}_{\vec{z}} = (\vec{W}^\trans\vec{\Sigma}^{-1}\vec{W} + \vec{\ident})^{-1}
\end{equation}
and 
\begin{equation}
	\hat{\vec{\mu}}_{\vec{z}} = \hat{\vec{\Sigma}}_{\vec{z}}\vec{W}^\trans\vec{\Sigma}^{-1}\vec{x}_n
\end{equation}

\newpage
\section{Bayesian CCA - Variational Bayes Derivation}\label{A: VI}

This appendix provides a full derivation for the update equations provided in Algorithm \ref{al: BayCCA VI algorithm}. These update equations differ with those presented by Wang \cite{wang2007} in that these equations assume independent columns of $\vec{W}$, matching the definition in \cite{klami2007}, rather than independent rows. Either definition (row or column) is suitable but one common approach was chosen here. Unlike \cite{wang2007}, this derivation also includes generic priors on the model parameters.

Given the model defined in Figure \ref{fig: bayesian cca graphical model}, the joint likelihood of Bayesian CCA can be defined as

\begin{equation}
	\mathcal{L} = p(\vec{x}, \vec{z}, \vec{W}, \vec{\Sigma}, \vec{\mu}) = p(\vec{x} | \vec{z}, \vec{W}, \vec{\Sigma}, \vec{\mu}) p(\vec{z}) p(\vec{\Sigma}) p(\vec{\mu}) p(\vec{W})
\end{equation}

Assuming independent columns of W, as stated in the model, and using precision $\mat{\Psi}$, the joint likelihood can be written as: 
\begin{equation}
	\mathcal{L} = p(\vec{x} | \vec{z}, \vec{W}, \vec{\Psi}, \vec{\mu}) p(\vec{z}) p(\vec{\Psi}) p(\vec{\mu}) \prod_{i=1}^{d} p(\vec{w}_i) 
\end{equation}
with the same components of the log likelihood as shown in the \ref{A: gibbs} derivation.  

The general premise of VI is to select a suitable approximation from a tractable family of distributions and try to make the approximation as close to the true (intractable) posterior as possible, usually by minimising the KL divergence. Assuming the surrogate posterior is determined by some free parameters, this problem reduces the inference to an optimisation problem. 

The following subsections summarise the derivation of the individual parameter updates needed to perform this optimisation. Even though a specified order is given in Algorithm \ref{al: BayCCA VI algorithm}, the VI scheme means updating the parameters can be conducted in any order, as long as the order remains fixed.  

Using a mean field approximation, the surrogate posterior is assumed to take the following factorised form
\begin{equation}
	q(\vec{z}, \vec{W}, \vec{\Psi}, \vec{\mu}) = q(\vec{z})  q(\vec{\Psi}) q(\vec{\mu}) \prod_{i=1}^{d} q(\vec{w}_i)
\end{equation}

where 

\begin{align}
    q(\vec{z}_n) \ \ &\sim \ \ \gaussdistr{\vec{z}_n | \breve{\vec{\mu}}_{\vec{z}}}{\breve{\vec{\Sigma}}_{\vec{z}}} \\
    q(\vec{\Psi}) \ \ &\sim \ \ \wishdistr{\vec{\Psi} | \breve{\vec{K}}^{-1}}{\breve{\nu}} \\
    q(\vec{w}_i) \ \ &\sim \ \ \gaussdistr{\vec{w}_i | \breve{\vec{\mu}}_{\vec{w}_i}}{\breve{\vec{\Sigma}}_{\vec{w}_i}} \\
    q(\vec{\mu}) \ \ &\sim \ \ \gaussdistr{\vec{\mu} | \breve{\vec{\mu}}_{\vec{\mu}}}{\breve{\vec{\Sigma}}_{\vec{\mu}}}
\end{align}

Given this factorised form, coordinate ascent VI (CAVI) \cite{murphy2012} updates for the parameters of the model can be found using 

\begin{equation}
	q^{\star}(\phi_k)= \expect[\phi \_]{\mathcal{L}}  = \expect[\phi \_]{p(\vec{x},\vec{z},\vec{\theta})} 
\end{equation}

where $\phi = \{\vec{z}, \vec{\theta}\}$, $\vec{\theta} = \{\vec{\Psi},\vec{\mu},\vec{W}\}$, $\phi \_$ denotes all elements of $\phi$ except the $k$th element being updated and $q^{\star}(\phi_k)$ refers to the updated surrogate posterior.

Working with the log likelihood, this can also be expressed as

\begin{equation}
	\ln{q^{\star}(\phi_k)} = \expect[\phi \_]{\ln{p(\vec{x},\vec{z},\vec{\theta})}} + \mathrm{const}\label{eq: q update general}
\end{equation}



\subsection{Update $\vec{z}$}

Collecting the terms of the log joint pertaining to $\vec{z}$ and substituting into Equation \ref{eq: q update general},

\begin{equation}
	\ln{q^{\star}(\vec{z}_n)} \propto \expect[\phi \_]{- \hlf \sum_{n=1}^N \left\{(\vec{x}_n - (\vec{W}\vec{z}_n + \vec{\mu}))^\trans \vec{\Psi} (\vec{x}_n - (\vec{W}\vec{z}_n + \vec{\mu})) \right\} - \hlf \sum_{n=1}^N\vec{z}_n^\trans \vec{z}_n}
\end{equation}

The expectation $\expect[\phi\_]{\bullet}$ is rewritten using $\left\langle \bullet \right\rangle _{\phi \_}$, such that

\begin{equation}
	\ln{q^{\star}(\vec{z}_n)} \propto \left\langle - \hlf \sum_{n=1}^N \left\{(\vec{x}_n - (\vec{W}\vec{z}_n + \vec{\mu}))^\trans \vec{\Psi} (\vec{x}_n - (\vec{W}\vec{z}_n + \vec{\mu})) + \vec{z}_n^\trans \vec{z}_n \right\} \right\rangle  _{\mat{W}, \ \vec{\mu},\ \mat{\Psi}}
\end{equation}

The subscript notation for the expectation is written once at the start of each update derivation with the relevant parameters and then omitted for clarity. 




Expanding and ignoring terms not containing $\vec{z}_n$, similar to the Gibbs sampling derivation, this reduces to


\begin{equation}
	\ln{q^{\star}(\vec{z}_n)} \propto \left\langle - \hlf\sum_{n=1}^N \left \{ \vec{z}_n^\trans \left(\vec{W}^\trans\vec{\Psi}\vec{W} + \vec{\ident}\right)\vec{z}_n - (\vec{x}_n - \vec{\mu})^\trans\vec{\Psi}\vec{W}\vec{z}_n - \vec{z}_n^\trans\vec{W}^\trans\vec{\Psi}(\vec{x}_n - \vec{\mu}) \right \} \right\rangle
\end{equation}


Setting $\breve{\vec{\Sigma}}^{-1}_{\vec{z}} = \langle \vec{W}^\trans \vec{\Psi}\vec{W}\rangle  + \vec{\ident}$, and given $\ident = \breve{\vec{\Sigma}}^{-1}_{\vec{z}}\breve{\vec{\Sigma}}_{\vec{z}}$, the log form of a Gaussian can be reached
\begin{equation}
	\ln{q^{\star}(\vec{z}_n)} \propto - \hlf\sum_{n=1}^N \left \{ \vec{z}_n^\trans \breve{\vec{\Sigma}}_{\vec{z}} \vec{z}_n - (\vec{x}_n - \langle \vec{\mu}\rangle )^\trans\langle \vec{\Psi}\rangle \langle \vec{W}\rangle \breve{\vec{\Sigma}}_{\vec{z}}\breve{\vec{\Sigma}}^{-1}_{\vec{z}} \vec{z}_n - \vec{z}_n^\trans \breve{\vec{\Sigma}}^{-1}_{\vec{z}} \breve{\vec{\Sigma}}_{\vec{z}} \langle \vec{W}\rangle ^\trans\langle \vec{\Psi}\rangle (\vec{x}_n - \langle \vec{\mu}\rangle ) \right \}
\end{equation}

Exponentiating, the surrogate posterior of $\vec{z}$ therefore has the following Gaussian form 

\begin{equation}
	q^{\star}(\vec{z}_n) \sim \gaussdistr{\breve{\vec{\mu}}_{\vec{z}}}{\breve{\vec{\Sigma}}_{\vec{z}}} 
\end{equation}
where 
\begin{equation}
	\breve{\vec{\Sigma}}_{\vec{z}} = \left(\langle \vec{W}^\trans \vec{\Psi}\vec{W}\rangle  + \vec{\ident} \right)^{-1}
\end{equation}
and 
\begin{equation}
	\breve{\vec{\mu}}_{\vec{z}} = \breve{\vec{\Sigma}}_{\vec{z}} \langle \vec{W}\rangle ^\trans\langle \vec{\Psi}\rangle (\vec{x}_n - \langle \vec{\mu}\rangle )
\end{equation}

\subsubsection*{Update $\vec{\Psi}$}   

Collecting the terms of the log joint pertaining to $\vec{\Psi}$ and substituting into Equation \ref{eq: q update general},

\begin{multline}
	\ln{q^{\star}(\vec{\Psi})} \propto \left\langle - \hlf \sum_{n=1}^N \left \{ \ln(|\vec{\Psi}|^{-1}) + (\vec{x} - (\mat{W}\vec{z}_n + \vec{\mu}))^{\trans}\mat{\Psi}(\vec{x} - (\mat{W}\vec{z}_n + \vec{\mu}))\right \} \right .  \\ \left . + \dfrac{\nu_0 - D + 1}{2}\ln{|\vec{\Psi}|} -  \hlf \mathrm{tr}(\vec{\Psi}\vec{K}_0) \right\rangle _{\mat{W}, \ \vec{\mu}, \ \vec{z}_n}
\end{multline}



Simplifying and rearranging terms, a log Wishart form can be reached such that

\begin{multline}
	\ln{q^{\star}(\vec{\Psi})} \propto  - \hlf \mathrm{tr} \left (\mat{\Psi} \left(\vec{K}_0 + \sum_{n=1}^N \left\langle (\vec{x} - (\mat{W}\vec{z}_n + \vec{\mu}))(\vec{x} - (\mat{W}\vec{z}_n + \vec{\mu}))^{\trans} \right\rangle \right) \right )  \\ + \dfrac{\nu_0 - D + 1 + N}{2}\ln{|\vec{\Psi}|}
\end{multline}

Exponentiating, the surrogate posterior of $\vec{\Psi}$ is then defined by, 

\begin{equation}
	q^{\star}(\vec{\Psi}) \sim \wishdistr{\breve{\vec{K}}^{-1}}{\breve{\nu}} 
\end{equation}
where 
\begin{equation}
	\breve{\vec{K}} = \vec{K}_0 +  \sum_{n=1}^N \left\langle (\vec{x}_n - \vec{\mu} - \vec{W}\vec{z}_n ) (\vec{x}_n - \vec{\mu} - \vec{W}\vec{z}_n)^\trans \right\rangle 
\end{equation} 
and 
\begin{equation}
	\breve{\nu} = \nu_0 + N
\end{equation}

\subsubsection*{Update $\vec{\mu}$}

Collecting the terms of the log joint pertaining to $\vec{\mu}$ and substituting into Equation \ref{eq: q update general},

\begin{multline}
	\ln{q^{\star}(\vec{\mu})} \propto \left\langle - \hlf \sum_{n=1}^N \left \{ (\vec{x} - (\mat{W}\vec{z}_n + \vec{\mu}))^{\trans}\mat{\Psi}(\vec{x} - (\mat{W}\vec{z}_n + \vec{\mu}))\right \} - \hlf \vec{\mu}^{\trans}\mat{\Sigma}_{\vec{\mu}}^{-1}\vec{\mu} \right .  \\ \left .- \hlf \vec{\mu}_{\vec{\mu}}^{\trans}\mat{\Sigma}_{\vec{\mu}}^{-1}\vec{\mu} - \hlf \vec{\mu}^{\trans}\mat{\Sigma}_{\vec{\mu}}^{-1}\vec{\mu}_{\vec{\mu}} \right\rangle _{\mat{W},\ \vec{z}_n,\ \mat{\Psi}}
\end{multline}



Simplifying,

\begin{multline}
	\ln{q^{\star}(\vec{\mu})} \propto - \hlf \left \{ \vec{\mu}^\trans (N\left\langle \vec{\Psi}\right\rangle + \mat{\Sigma}_{\vec{\mu}}^{-1}) \vec{\mu} - \vec{\mu}^\trans \sum_{n=1}^N\left\langle \vec{\Psi} \right\rangle (\vec{x}_n - \left\langle \vec{W} \right\rangle \left\langle\vec{z}_n \right\rangle ) \right.  \\ \left. - \sum_{n=1}^N(\vec{x}_n - \left\langle \vec{W} \right\rangle \left\langle\vec{z}_n \right\rangle )^\trans \left\langle \vec{\Psi}\right\rangle \vec{\mu}  + \vec{\mu}_{\vec{\mu}}^{\trans}\mat{\Sigma}_{\vec{\mu}}^{-1}\vec{\mu} + \vec{\mu}^{\trans}\mat{\Sigma}_{\vec{\mu}}^{-1}\vec{\mu}_{\vec{\mu}} \right \}
\end{multline}

letting $\breve{\vec{\Sigma}}^{-1}_{\mu} = N\left\langle \vec{\Psi}\right\rangle + \mat{\Sigma}_{\vec{\mu}}^{-1}$, and given $\ident = \breve{\vec{\Sigma}}^{-1}_{\mu}\breve{\vec{\Sigma}}_{\mu}$,

\begin{multline}
	\ln{q^{\star}(\vec{\mu})} \propto - \hlf \left \{ \vec{\mu}^\trans \breve{\vec{\Sigma}}^{-1}_{\mu} \vec{\mu} - \vec{\mu}^\trans \breve{\vec{\Sigma}}^{-1}_{\mu}\breve{\vec{\Sigma}}_{\mu} \left (\sum_{n=1}^N\left\langle \vec{\Psi} \right\rangle (\vec{x}_n - \left\langle \vec{W} \right\rangle \left\langle\vec{z}_n \right\rangle ) + \mat{\Sigma}_{\vec{\mu}}^{-1}\vec{\mu}_{\vec{\mu}} \right ) \right.  \\ \left. - \left ( \sum_{n=1}^N(\vec{x}_n - \left\langle \vec{W} \right\rangle \left\langle\vec{z}_n \right\rangle )^\trans \left\langle \vec{\Psi}\right\rangle + \vec{\mu}_{\vec{\mu}}^{\trans}\mat{\Sigma}_{\vec{\mu}}^{-1} \right )\breve{\vec{\Sigma}}_{\mu}\breve{\vec{\Sigma}}^{-1}_{\mu}\vec{\mu}  \right \}
\end{multline}

Exponentiating, the surrogate posterior of $\vec{\mu}$ thus has the following form 

\begin{equation}
	q^{\star}(\vec{\mu}) \sim \gaussdistr{\breve{\vec{\mu}}_{\vec{\mu}}}{\breve{\vec{\Sigma}}_{\vec{\mu}}} 
\end{equation}
where 
\begin{equation}
	\breve{\vec{\Sigma}}_{\vec{\mu}} = (N\langle\vec{\Psi}\rangle + \vec{\Sigma}_{\mu}^{-1})^{-1}
\end{equation}
and 
\begin{equation}
	\breve{\vec{\mu}}_{\vec{\mu}} = \breve{\vec{\Sigma}}_{\mu}\left(\langle\vec{\Psi}\rangle \sum_{n=1}^N (\vec{x}_n - \langle\vec{W}\rangle\langle\vec{z}_n\rangle) + \vec{\Sigma}_{\mu}^{-1}\vec{\mu}_{\mu}\right)
\end{equation}

\subsubsection*{Update $\vec{w}_i$}

Collecting the terms of the log joint pertaining to $\vec{w}_i$ and substituting into Equation \ref{eq: q update general},

\begin{multline}
	\ln{q^{\star}(\vec{w}_i)} \propto \left\langle - \hlf \sum_{n=1}^N \left \{ (\vec{x} - (\mat{W}\vec{z}_n + \vec{\mu}))^{\trans}\mat{\Psi}(\vec{x} - (\mat{W}\vec{z}_n + \vec{\mu}))\right \} - \hlf \vec{w}_i^{\trans}\mat{\Sigma}_{\vec{w}_i}^{-1}\vec{w}_i \right .  \\ \left .- \hlf \vec{\mu}_{\vec{w}_i}^{\trans}\mat{\Sigma}_{\vec{w}_i}^{-1}\vec{w}_i - \hlf \vec{w}_i^{\trans}\mat{\Sigma}_{\vec{w}_i}^{-1}\vec{\mu}_{\vec{w}_i} \right\rangle _{\vec{\mu},\ \vec{z}_n,\ \mat{\Psi}} \label{eq: log joint wi}
\end{multline}

To construct the update for a single column, the full weight matrix must be separated into the column of interest and the remaining columns. This is achieved by defining

\begin{equation}
	\tilde{\vec{x}}_n  = \vec{x}_n - \vec{\mu} - \vec{w}_{\_i}\vec{z}_{\_i,n}
\end{equation}

where $\vec{w}_{\_i}$ corresponds to all columns except the $i$th column of interest and $\vec{z}_{\_i,n}$ corresponds to all rows of $\vec{z}_n$ except the $i$th row. Using this definition, Equation \ref{eq: log joint wi} becomes

\begin{multline}
	\ln{q^{\star}(\vec{w}_i)} \propto \left\langle - \hlf \sum_{n=1}^N \left \{ (\tilde{\vec{x}}_n - \vec{w}_i \vec{z}_{i,n})^{\trans}\mat{\Psi}(\tilde{\vec{x}}_n - \vec{w}_i \vec{z}_{i,n})\right \} - \hlf \vec{w}_i^{\trans}\mat{\Sigma}_{\vec{w}_i}^{-1}\vec{w}_i \right .  \\ \left .- \hlf \vec{\mu}_{\vec{w}_i}^{\trans}\mat{\Sigma}_{\vec{w}_i}^{-1}\vec{w}_i - \hlf \vec{w}_i^{\trans}\mat{\Sigma}_{\vec{w}_i}^{-1}\vec{\mu}_{\vec{w}_i} \right\rangle
\end{multline}


As $\vec{z}_{i,n}$ is scalar, it can be manipulated as one. Rearranging and combining terms, this gives
\begin{multline}
	\ln{q^{\star}(\vec{w}_i)} \propto - \hlf \left\langle \vec{w}_i^{\trans}\left (\mat{\Sigma}_{\vec{w}_i}^{-1}  + \sum_{n=1}^N \vec{z}_{i,n}\vec{z}_{i,n} \vec{\Psi} \right )\vec{w}_i -  \vec{w}_i^{\trans} \left (\sum_{n=1}^N \vec{z}_{i,n}\vec{\Psi}\tilde{\vec{x}}_n + \mat{\Sigma}_{\vec{w}_i}^{-1}\vec{\mu}_{\vec{w}_i} \right )  \right .  \\ \left . - \left( \sum_{n=1}^N\tilde{\vec{x}}_n^{\trans}\mat{\Psi}\vec{z}_{i,n} + \vec{\mu}_{\vec{w}_i}^{\trans}\mat{\Sigma}_{\vec{w}_i}^{-1} \right) \vec{w}_i \right\rangle
\end{multline}

Considering the expectations, 

\begin{multline}
	\ln{q^{\star}(\vec{w}_i)} \propto - \hlf \left \{ \vec{w}_i^{\trans} \left (\mat{\Sigma}_{\vec{w}_i}^{-1}  + \sum_{n=1}^N \left\langle\vec{z}_{i,n}\vec{z}_{i,n} \right\rangle \left\langle \vec{\Psi} \right\rangle \right )\vec{w}_i -  \vec{w}_i^{\trans} \left (\sum_{n=1}^N \left\langle\vec{z}_{i,n}\right\rangle \left\langle\vec{\Psi}\right\rangle \tilde{\vec{x}}_n + \mat{\Sigma}_{\vec{w}_i}^{-1}\vec{\mu}_{\vec{w}_i} \right ) \right.  \\ \left. - \left( \sum_{n=1}^N\tilde{\vec{x}}_n^{\trans} \left\langle\mat{\Psi}\right\rangle \left\langle\vec{z}_{i,n}\right\rangle + \vec{\mu}_{\vec{w}_i}^{\trans}\mat{\Sigma}_{\vec{w}_i}^{-1} \right) \vec{w}_i \right \}
\end{multline}

letting $\breve{\vec{\Sigma}}^{-1}_{\vec{w}_i} = \left (\mat{\Sigma}_{\vec{w}_i}^{-1}  + \sum_{n=1}^N \left\langle\vec{z}_{i,n}\vec{z}_{i,n} \right\rangle \left\langle \vec{\Psi} \right\rangle \right )$, and as $\ident = \breve{\vec{\Sigma}}^{-1}_{\vec{w}_i}\breve{\vec{\Sigma}}_{\vec{w}_i}$,

\begin{equation}
	q^{\star}(\vec{w}_i) \sim \gaussdistr{\breve{\vec{\mu}}_{\vec{w}_i}}{\breve{\vec{\Sigma}}_{\vec{w}_i}} 
\end{equation}
where 
\begin{equation}
	\breve{\vec{\Sigma}}_{\vec{w}_i} = (\sum^{N}_{n=1} \langle \vec{z}_{i,n}\vec{z}_{i,n} \rangle \langle\vec{\Psi}\rangle + \vec{\Sigma}_{\vec{w}_i}^{-1})^{-1}
\end{equation}
and 
\begin{equation}
	\breve{\vec{\mu}}_{\vec{w}_i} = \breve{\vec{\Sigma}}_{\vec{w}_i}\left(\langle\vec{\Psi}\rangle \sum_{n=1}^N \tilde{\vec{x}}_n\langle\vec{z}_{n,i}^{\trans}\rangle) + \vec{\Sigma}_{\vec{w}_i}^{-1}\vec{\mu}_{\vec{w}_i}\right)
\end{equation}

\newpage

\bibliographystyle{unsrtnatdrg}
\bibliography{references2.bib}

\begin{thebibliography}{67}
\providecommand{\natexlab}[1]{#1}
\providecommand{\url}[1]{\texttt{#1}}
\expandafter\ifx\csname urlstyle\endcsname\relax
  \providecommand{\doi}[1]{doi: #1}\else
  \providecommand{\doi}{doi: \begingroup \urlstyle{rm}\Url}\fi

\bibitem[Brincker and Ventura(2015)]{brincker2015}
R~Brincker and C.~E Ventura.
\newblock \emph{Introduction to {{Operational Modal Analysis}}}.
\newblock John Wiley \& Sons, Ltd, Chichester, UK, 2015.

\bibitem[Brincker et~al.(2001)Brincker, Zhang, and Andersen]{brincker2001}
R~Brincker, L~Zhang, and P~Andersen.
\newblock Modal identification of output-only systems using frequency domain
  decomposition.
\newblock \emph{Smart Materials and Structures}, 10\penalty0 (3):\penalty0
  441--445, 2001.

\bibitem[Van~Overschee and De~Moor(1996)]{vanoverschee1996}
P~Van~Overschee and B~De~Moor.
\newblock \emph{Subspace {{Identification}} for {{Linear Systems}}}.
\newblock Springer US, Boston, MA, 1996.

\bibitem[Katayama(2005)]{katayama2005}
T~Katayama.
\newblock \emph{Subspace Methods for System Identification: {{A}} Realisation
  Approach.}
\newblock Communications and Control Engineering. Springer, London, 2005.

\bibitem[Farrar and Worden(2013)]{farrar2013}
C.~R Farrar and K~Worden.
\newblock \emph{Structural Health Monitoring: A Machine Learning Perspective}.
\newblock Wiley, Chichester, 2013.

\bibitem[Mclean et~al.(2023)Mclean, Jones, O'Connell, Maguire, and
  Rogers]{mclean2023}
J.~H Mclean, M.~R Jones, B.~J O'Connell, E~Maguire, and T.~J Rogers.
\newblock Physically meaningful uncertainty quantification in probabilistic
  wind turbine power curve models as a damage-sensitive feature.
\newblock \emph{Structural Health Monitoring}, 22\penalty0 (6):\penalty0
  3623--3636, 2023.

\bibitem[Li(2016)]{li2016}
B~Li.
\newblock \emph{Uncertainty {{Quantification}} in {{Vibration-based Structural
  Health Monitoring}} Using {{Bayesian Statistics}}}.
\newblock PhD thesis, University of California, Berkeley, 2016.

\bibitem[Bull et~al.(2021)Bull, Gardner, Rogers, Cross, Dervilis, and
  Worden]{bull2021}
L.~A Bull, P~Gardner, T.~J Rogers, E.~J Cross, N~Dervilis, and K~Worden.
\newblock Probabilistic {{Inference}} for {{Structural Health Monitoring}}:
  {{New Modes}} of {{Learning}} from {{Data}}.
\newblock \emph{ASCE-ASME Journal of Risk and Uncertainty in Engineering
  Systems, Part A: Civil Engineering}, 7\penalty0 (1):\penalty0 03120003, 2021.

\bibitem[Miah and Lienhart(2022)]{miah2022}
M~Miah and W~Lienhart.
\newblock Uncertainties of {{Parameters Quantification}} in {{SHM}}.
\newblock In \emph{8th {{European Congress}} on {{Computational Methods}} in
  {{Applied Sciences}} and {{Engineering}}}. CIMNE, 2022.

\bibitem[Wagg et~al.(2020)Wagg, Worden, Barthorpe, and Gardner]{wagg2020a}
D.~J Wagg, K~Worden, R.~J Barthorpe, and P~Gardner.
\newblock Digital {{Twins}}: {{State-of-the-Art}} and {{Future Directions}} for
  {{Modeling}} and {{Simulation}} in {{Engineering Dynamics Applications}}.
\newblock \emph{ASCE-ASME Journal of Risk and Uncertainty in Engineering
  Systems, Part B: Mechanical Engineering}, 6\penalty0 (3):\penalty0 030901,
  2020.

\bibitem[Thelen et~al.(2023)Thelen, Zhang, Fink, Lu, Ghosh, Youn, Todd,
  Mahadevan, Hu, and Hu]{thelen2023}
A~Thelen, X~Zhang, O~Fink, Y~Lu, S~Ghosh, B.~D Youn, M.~D Todd, S~Mahadevan,
  C~Hu, and Z~Hu.
\newblock A comprehensive review of digital twin---part 2: Roles of uncertainty
  quantification and optimization, a battery digital twin, and perspectives.
\newblock \emph{Structural and Multidisciplinary Optimization}, 66\penalty0
  (1):\penalty0 1, 2023.

\bibitem[R{\'i}os et~al.(2020)R{\'i}os, Staudter, Weber, Anderl, and
  Bernard]{rios2020}
J~R{\'i}os, G~Staudter, M~Weber, R~Anderl, and A~Bernard.
\newblock Uncertainty of data and the digital twin: A review.
\newblock \emph{International Journal of Product Lifecycle Management},
  12\penalty0 (4):\penalty0 329--358, 2020.

\bibitem[Kochunas and Huan(2021)]{kochunas2021}
B~Kochunas and X~Huan.
\newblock Digital {{Twin Concepts}} with {{Uncertainty}} for {{Nuclear Power
  Applications}}.
\newblock \emph{Energies}, 14\penalty0 (14):\penalty0 4235, 2021.

\bibitem[Hughes(2022)]{hughes2022}
A.~J Hughes.
\newblock \emph{On {{Risk-Based Decision-Making}} for {{Structural Health
  Monitoring}}}.
\newblock PhD thesis, University of Sheffield, Sheffield, UK, 2022.

\bibitem[Ciloglu et~al.(2012)Ciloglu, Zhou, Moon, and Aktan]{ciloglu2012}
K~Ciloglu, Y~Zhou, F~Moon, and A.~E Aktan.
\newblock Impacts of {{Epistemic Uncertainty}} in {{Operational Modal
  Analysis}}.
\newblock \emph{Journal of Engineering Mechanics}, 138\penalty0 (9):\penalty0
  1059--1070, 2012.

\bibitem[Reynders et~al.(2008)Reynders, Pintelon, and De~Roeck]{reynders2008}
E~Reynders, R~Pintelon, and G~De~Roeck.
\newblock Uncertainty bounds on modal parameters obtained from stochastic
  subspace identification.
\newblock \emph{Mechanical Systems and Signal Processing}, 22\penalty0
  (4):\penalty0 948--969, 2008.

\bibitem[Yang and Lam(2019)]{yang2019}
J.-H Yang and H.-F Lam.
\newblock An innovative {{Bayesian}} system identification method using
  autoregressive model.
\newblock \emph{Mechanical Systems and Signal Processing}, 133:\penalty0
  106289, 2019.

\bibitem[Kang and Zeng(2023)]{kang2023}
J~Kang and S~Zeng.
\newblock Uncertainty quantification in operational modal analysis of
  time-varying structures based on time-dependent autoregressive moving average
  model.
\newblock \emph{Journal of Sound and Vibration}, 548:\penalty0 117549, 2023.

\bibitem[Rogers et~al.(2020)Rogers, Worden, and Cross]{rogers2020a}
T.~J Rogers, K~Worden, and E.~J Cross.
\newblock On the application of {{Gaussian}} process latent force models for
  joint input-state-parameter estimation: {{With}} a view to {{Bayesian}}
  operational identification.
\newblock \emph{Mechanical Systems and Signal Processing}, 140:\penalty0
  106580, 2020.

\bibitem[Au(2012{\natexlab{a}})]{au2012}
S.-K Au.
\newblock Fast {{Bayesian}} ambient modal identification in the frequency
  domain, {{Part I}}: {{Posterior}} most probable value.
\newblock \emph{Mechanical Systems and Signal Processing}, 26:\penalty0 60--75,
  2012{\natexlab{a}}.

\bibitem[Au(2012{\natexlab{b}})]{au2012a}
S.-K Au.
\newblock Fast {{Bayesian}} ambient modal identification in the frequency
  domain, {{Part II}}: {{Posterior}} uncertainty.
\newblock \emph{Mechanical Systems and Signal Processing}, 26:\penalty0 76--90,
  2012{\natexlab{b}}.

\bibitem[Beck(2010)]{beck2010a}
J.~L Beck.
\newblock Bayesian system identification based on probability logic.
\newblock \emph{Structural Control and Health Monitoring}, 17\penalty0
  (7):\penalty0 825--847, 2010.

\bibitem[Huang et~al.(2017)Huang, Beck, and Li]{huang2017}
Y~Huang, J.~L Beck, and H~Li.
\newblock Bayesian system identification based on hierarchical sparse
  {{Bayesian}} learning and {{Gibbs}} sampling with application to structural
  damage assessment.
\newblock \emph{Computer Methods in Applied Mechanics and Engineering},
  318:\penalty0 382--411, 2017.

\bibitem[Reynders et~al.(2016)Reynders, Maes, Lombaert, and
  De~Roeck]{reynders2016}
E~Reynders, K~Maes, G~Lombaert, and G~De~Roeck.
\newblock Uncertainty quantification in operational modal analysis with
  stochastic subspace identification: {{Validation}} and applications.
\newblock \emph{Mechanical Systems and Signal Processing}, 66--67:\penalty0
  13--30, 2016.

\bibitem[Reynders(2021)]{reynders2021}
E.~P Reynders.
\newblock Uncertainty quantification in data-driven stochastic subspace
  identification.
\newblock \emph{Mechanical Systems and Signal Processing}, 151:\penalty0
  107338, 2021.

\bibitem[D{\"o}hler and Mevel(2013)]{dohler2013a}
M~D{\"o}hler and L~Mevel.
\newblock Efficient multi-order uncertainty computation for stochastic subspace
  identification.
\newblock \emph{Mechanical Systems and Signal Processing}, 38\penalty0
  (2):\penalty0 346--366, 2013.

\bibitem[D{\"o}hler et~al.(2013)D{\"o}hler, Lam, and Mevel]{dohler2013}
M~D{\"o}hler, X.-B Lam, and L~Mevel.
\newblock Uncertainty quantification for modal parameters from stochastic
  subspace identification on multi-setup measurements.
\newblock \emph{Mechanical Systems and Signal Processing}, 36\penalty0
  (2):\penalty0 562--581, 2013.

\bibitem[Gres and D{\"o}hler(2022)]{gres2022}
S~Gres and M~D{\"o}hler.
\newblock Uncertainty propagation in subspace methods for operational modal
  analysis under misspecified model orders.
\newblock In \emph{{{ISMA-USD}} 2022 - {{International Conference}} on
  {{Noise}} and {{Vibration Engineering}}, and {{Uncertainty Structural
  Dynamics}}}, Leuven, Belgium, 2022.

\bibitem[Gre{\'s} et~al.(2022)Gre{\'s}, Riva, S{\"u}leyman, Andersen, and
  {\L}uczak]{gres2022a}
S~Gre{\'s}, R~Riva, C.~Y S{\"u}leyman, P~Andersen, and M.~M {\L}uczak.
\newblock Uncertainty quantification of modal parameter estimates obtained from
  subspace identification: {{An}} experimental validation on a laboratory test
  of a large-scale wind turbine blade.
\newblock \emph{Engineering Structures}, 256:\penalty0 114001, 2022.

\bibitem[Pintelon et~al.(2007)Pintelon, Guillaume, and Schoukens]{pintelon2007}
R~Pintelon, P~Guillaume, and J~Schoukens.
\newblock Uncertainty calculation in (operational) modal analysis.
\newblock \emph{Mechanical Systems and Signal Processing}, 21\penalty0
  (6):\penalty0 2359--2373, 2007.

\bibitem[{El-kafafy} et~al.(2013){El-kafafy}, De~Troyer, Peeters, and
  Guillaume]{el-kafafy2013a}
M~{El-kafafy}, T~De~Troyer, B~Peeters, and P~Guillaume.
\newblock Fast maximum-likelihood identification of modal parameters with
  uncertainty intervals: {{A}} modal model-based formulation.
\newblock \emph{Mechanical Systems and Signal Processing}, 37\penalty0
  (1-2):\penalty0 422--439, 2013.

\bibitem[Mellinger et~al.(2016)Mellinger, D{\"o}hler, and Mevel]{mellinger2016}
P~Mellinger, M~D{\"o}hler, and L~Mevel.
\newblock Variance estimation of modal parameters from output-only and
  input/output subspace-based system identification.
\newblock \emph{Journal of Sound and Vibration}, 379:\penalty0 1--27, 2016.

\bibitem[Reynders(2012)]{reynders2012}
E~Reynders.
\newblock System {{Identification Methods}} for ({{Operational}}) {{Modal
  Analysis}}: {{Review}} and {{Comparison}}.
\newblock \emph{Archives of Computational Methods in Engineering}, 19\penalty0
  (1):\penalty0 51--124, 2012.

\bibitem[Pereira et~al.(2020)Pereira, Reynders, Magalh{\~a}es, Cunha, and
  Gomes]{pereira2020}
S~Pereira, E~Reynders, F~Magalh{\~a}es, {\'A}~Cunha, and J.~P Gomes.
\newblock The role of modal parameters uncertainty estimation in automated
  modal identification, modal tracking and data normalization.
\newblock \emph{Engineering Structures}, 224:\penalty0 111208, 2020.

\bibitem[Priou et~al.(2022)Priou, Gres, Perrault, Guerineau, and
  D{\"o}hler]{priou2022}
J~Priou, S~Gres, M~Perrault, L~Guerineau, and M~D{\"o}hler.
\newblock Automated uncertainty-based extraction of modal parameters from
  stabilization diagrams.
\newblock In \emph{9th {{International Operational Modal Analysis
  Conference}}}, Vancouver, BC, Canada, 2022.

\bibitem[Gre{\'s} and D{\"o}hler(2023)]{gres2023}
S~Gre{\'s} and M~D{\"o}hler.
\newblock Model {{Order Selection}} for {{Uncertainty Quantification}} in
  {{Subspace-Based OMA}} of {{Vestas V27 Blade}}.
\newblock In M.~P Limongelli, P.~F Giordano, S~Quqa, C~Gentile, and A~Cigada,
  editors, \emph{Experimental {{Vibration Analysis}} for {{Civil Engineering
  Structures}}}, volume 433, pages 43--52. Springer Nature Switzerland, Cham,
  2023.

\bibitem[Au et~al.(2013)Au, Zhang, and Ni]{au2013}
S.-K Au, F.-L Zhang, and Y.-C Ni.
\newblock Bayesian operational modal analysis: {{Theory}}, computation,
  practice.
\newblock \emph{Computers \& Structures}, 126:\penalty0 3--14, 2013.

\bibitem[Yuen et~al.(2002)Yuen, Katafygiotis, and Beck]{yuen2002}
K.-V Yuen, L.~S Katafygiotis, and J.~L Beck.
\newblock Spectral density estimation of stochastic vector processes.
\newblock \emph{Probabilistic Engineering Mechanics}, 17\penalty0 (3):\penalty0
  265--272, 2002.

\bibitem[Yuen and Katafygiotis(2003)]{yuen2003}
K.-V Yuen and L.~S Katafygiotis.
\newblock Bayesian {{Fast Fourier Transform Approach}} for {{Modal Updating
  Using Ambient Data}}.
\newblock \emph{Advances in Structural Engineering}, 6\penalty0 (2):\penalty0
  81--95, 2003.

\bibitem[Au(2012{\natexlab{c}})]{au2012b}
S.-K Au.
\newblock Connecting {{Bayesian}} and frequentist quantification of parameter
  uncertainty in system identification.
\newblock \emph{Mechanical Systems and Signal Processing}, 29:\penalty0
  328--342, 2012{\natexlab{c}}.

\bibitem[Au(2016)]{au2016}
S.-K Au.
\newblock Insights on the {{Bayesian}} spectral density method for operational
  modal analysis.
\newblock \emph{Mechanical Systems and Signal Processing}, 66--67:\penalty0
  1--12, 2016.

\bibitem[Au(2017)]{au2017}
S.-K Au.
\newblock \emph{Operational {{Modal Analysis}}}.
\newblock Springer Singapore, Singapore, 2017.

\bibitem[Au et~al.(2018)Au, Brownjohn, and Mottershead]{au2018}
S.-K Au, J.~M Brownjohn, and J.~E Mottershead.
\newblock Quantifying and managing uncertainty in operational modal analysis.
\newblock \emph{Mechanical Systems and Signal Processing}, 102:\penalty0
  139--157, 2018.

\bibitem[Zhu and Au(2018{\natexlab{a}})]{zhu2018}
Y.-C Zhu and S.-K Au.
\newblock Bayesian operational modal analysis with asynchronous data, part
  {{I}}: {{Most}} probable value.
\newblock \emph{Mechanical Systems and Signal Processing}, 98:\penalty0
  652--666, 2018{\natexlab{a}}.

\bibitem[Zhu and Au(2018{\natexlab{b}})]{zhu2018a}
Y.-C Zhu and S.-K Au.
\newblock Bayesian operational modal analysis with asynchronous data, {{Part
  II}}: {{Posterior}} uncertainty.
\newblock \emph{Mechanical Systems and Signal Processing}, 98:\penalty0
  920--935, 2018{\natexlab{b}}.

\bibitem[Zhu and Au(2020)]{zhu2020}
Y.-C Zhu and S.-K Au.
\newblock Bayesian data driven model for uncertain modal properties identified
  from operational modal analysis.
\newblock \emph{Mechanical Systems and Signal Processing}, 136:\penalty0
  106511, 2020.

\bibitem[Zhu et~al.(2021)Zhu, Au, Li, and Xie]{zhu2021}
Z~Zhu, S.-K Au, B~Li, and Y.-L Xie.
\newblock Bayesian operational modal analysis with multiple setups and multiple
  (possibly close) modes.
\newblock \emph{Mechanical Systems and Signal Processing}, 150:\penalty0
  107261, 2021.

\bibitem[Mao et~al.(2023)Mao, Su, Wang, and Li]{mao2023}
J~Mao, X~Su, H~Wang, and J~Li.
\newblock Automated {{Bayesian}} operational modal analysis of the long-span
  bridge using machine-learning algorithms.
\newblock \emph{Engineering Structures}, 289:\penalty0 116336, 2023.

\bibitem[Brownjohn et~al.(2018)Brownjohn, Au, Zhu, Sun, Li, Bassitt, Hudson,
  and Sun]{brownjohn2018}
J.~M.~W Brownjohn, S.-K Au, Y~Zhu, Z~Sun, B~Li, J~Bassitt, E~Hudson, and H~Sun.
\newblock Bayesian operational modal analysis of {{Jiangyin Yangtze River
  Bridge}}.
\newblock \emph{Mechanical Systems and Signal Processing}, 110:\penalty0
  210--230, 2018.

\bibitem[Brownjohn et~al.(2019)Brownjohn, Raby, Au, Zhu, Wang, Antonini,
  Pappas, and D'Ayala]{brownjohn2019}
J.~M.~W Brownjohn, A~Raby, S.-K Au, Z~Zhu, X~Wang, A~Antonini, A~Pappas, and
  D~D'Ayala.
\newblock Bayesian operational modal analysis of offshore rock lighthouses:
  {{Close}} modes, alignment, symmetry and uncertainty.
\newblock \emph{Mechanical Systems and Signal Processing}, 133:\penalty0
  106306, 2019.

\bibitem[Yan and Katafygiotis(2015)]{yan2015}
W.-J Yan and L.~S Katafygiotis.
\newblock A two-stage fast {{Bayesian}} spectral density approach for ambient
  modal analysis. {{Part II}}: {{Mode}} shape assembly and case studies.
\newblock \emph{Mechanical Systems and Signal Processing}, 54--55:\penalty0
  156--171, 2015.

\bibitem[Li and Der~Kiureghian(2017)]{li2017}
B~Li and A~Der~Kiureghian.
\newblock Operational modal identification using variational {{Bayes}}.
\newblock \emph{Mechanical Systems and Signal Processing}, 88:\penalty0
  377--398, 2017.

\bibitem[Li et~al.(2018)Li, Der~Kiureghian, and Au]{li2018a}
B~Li, A~Der~Kiureghian, and S.-K Au.
\newblock A {{Gibbs}} sampling algorithm for structural modal identification
  under seismic excitation.
\newblock \emph{Earthquake Engineering \& Structural Dynamics}, 47\penalty0
  (14):\penalty0 2735--2755, 2018.

\bibitem[Dollon et~al.(2022)Dollon, Antoni, Tahan, Gagnon, and
  Monette]{dollon2022}
Q~Dollon, J~Antoni, A~Tahan, M~Gagnon, and C~Monette.
\newblock A fast collapsed {{Gibbs}} sampler for frequency domain operational
  modal analysis.
\newblock \emph{Mechanical Systems and Signal Processing}, 173:\penalty0
  108985, 2022.

\bibitem[Hotelling(1936)]{hotelling1936}
H~Hotelling.
\newblock Relations between two sets of variates.
\newblock \emph{Biometrika}, 28\penalty0 (3/4):\penalty0 321--375, 1936.

\bibitem[Bach and Jordan(2005)]{bach2005}
F.~R Bach and M.~I Jordan.
\newblock A {{Probabilistic Interpretation}} of {{Canonical Correlation
  Analysis}}.
\newblock Technical {{Report}} 688, University of California, Berkeley, 2005.

\bibitem[Koller and Friedman(2009)]{koller2009}
D~Koller and N~Friedman.
\newblock \emph{Probabilistic Graphical Models: Principles and Techniques}.
\newblock Adaptive Computation and Machine Learning. MIT Press, Cambridge, MA,
  2009.

\bibitem[O'Connell and Rogers(2024)]{oconnell2024}
B.~J O'Connell and T.~J Rogers.
\newblock A robust probabilistic approach to stochastic subspace
  identification.
\newblock \emph{Journal of Sound and Vibration}, 581:\penalty0 118381, 2024.

\bibitem[Wang(2007)]{wang2007}
C~Wang.
\newblock Variational {{Bayesian Approach}} to {{Canonical Correlation
  Analysis}}.
\newblock \emph{IEEE Transactions on Neural Networks}, 18\penalty0
  (3):\penalty0 905--910, 2007.

\bibitem[Klami and Kaski(2007)]{klami2007}
A~Klami and S~Kaski.
\newblock Local dependent components.
\newblock In \emph{Proceedings of the 24th International Conference on
  {{Machine}} Learning}, pages 425--432, Corvalis Oregon USA, 2007. ACM.

\bibitem[Murphy(2012)]{murphy2012}
K.~P Murphy.
\newblock \emph{Machine {{Learning}}: {{A Probabilistic Perspective}}}.
\newblock The MIT Press, 2012.

\bibitem[Bishop(2006)]{bishop2006}
C.~M Bishop.
\newblock \emph{Pattern Recognition and Machine Learning}.
\newblock Information Science and Statistics. Springer, New York, 2006.

\bibitem[Opper and Saad(2001)]{Opper2001}
M~Opper and D~Saad, editors.
\newblock \emph{Advanced mean field methods: theory and practice}.
\newblock Neural Information Processing. MIT, Cambridge, Massachusetts (US),
  2001.

\bibitem[S{\"a}rkk{\"a} and Solin(2019)]{sarkka2019a}
S~S{\"a}rkk{\"a} and A~Solin.
\newblock \emph{Applied {{Stochastic Differential Equations}}}.
\newblock Cambridge University Press, 2019.

\bibitem[Maeck and De~Roeck(2003)]{maeck2003}
J~Maeck and G~De~Roeck.
\newblock {{Description of Z24 benchmark}}.
\newblock \emph{Mechanical Systems and Signal Processing}, 17\penalty0
  (1):\penalty0 127--131, 2003.

\bibitem[Mevel et~al.(2003)Mevel, Goursat, and Basseville]{mevel2003}
L~Mevel, M~Goursat, and M~Basseville.
\newblock Stochastic supspace-based structureal identification and damage
  detection and localisation - application to the {{Z24}} bridge benchmark.
\newblock \emph{Mechanical Systems and Signal Processing}, 17\penalty0
  (1):\penalty0 143--151, 2003.

\bibitem[Ewins(2000)]{ewins2000}
D.~J Ewins.
\newblock \emph{Modal Testing: Theory, Practice and Application}.
\newblock Number~10 in Mechanical Engineering Research Studies Engineering
  Dynamics Series. Research Studies Press, Baldock, 2000.

\end{thebibliography}

\end{document}